\documentclass[10pt]{article}

\usepackage[letterpaper,margin=1in]{geometry}
\usepackage[T1]{fontenc}
\usepackage[utf8]{inputenc}
\usepackage{amsmath,amsthm}
\usepackage{newtxtext,newtxmath}
\usepackage{microtype}
\usepackage{bm}  %
\usepackage{graphicx}
\usepackage{booktabs,multirow,array,tabularx}
\usepackage[table]{xcolor}
\definecolor{oursrow}{RGB}{221,240,221}  %
\usepackage[labelfont=bf,font=small,labelsep=period]{caption}
\usepackage{subcaption}
\usepackage{enumitem}
\usepackage{titlesec}
\usepackage[numbers,square,sort&compress]{natbib}
\usepackage[colorlinks=true,linkcolor=blue!60!black,citecolor=blue!60!black,urlcolor=blue!60!black]{hyperref}
\usepackage[capitalise,noabbrev]{cleveref}
\usepackage{pifont}

\titleformat{\section}{\sffamily\bfseries\large}{\thesection}{0.6em}{}
\titleformat{\subsection}{\sffamily\bfseries\normalsize}{\thesubsection}{0.6em}{}
\titlespacing*{\section}{0pt}{1.4ex plus .5ex minus .2ex}{0.8ex}
\titlespacing*{\subsection}{0pt}{1.1ex plus .4ex minus .2ex}{0.5ex}
\titlespacing*{\paragraph}{0pt}{0.8ex plus .3ex}{0.8em}
\setlist{nosep,leftmargin=*}

\usepackage{tikz}

\usetikzlibrary{positioning,arrows.meta,calc,fit,backgrounds}
\definecolor{formulablue}{HTML}{DCEAF7}
\definecolor{learngrey}{HTML}{D9D9D9}
\definecolor{policygreen}{HTML}{DFF0D8}
\definecolor{markgreen}{RGB}{26,140,72}
\definecolor{markred}{RGB}{200,45,45}
\newcommand{\cmark}{\textcolor{markgreen}{\ding{51}}}
\newcommand{\xmark}{\textcolor{markred}{\ding{55}}}

\title{\textbf{Myodyssey: A Muscle-Driven Motion Generalist Trained\\through a Differentiable Musculoskeletal World Model}}
\author{Yifei Yuan \qquad Xianlian Zhou\\[2pt]
{\normalsize Department of Biomedical Engineering, New Jersey Institute of Technology, Newark, NJ 07102, USA}\\
{\normalsize \texttt{yy72@njit.edu}, \texttt{alexzhou@njit.edu}}\\[2pt]
{\normalsize Project page: \url{https://fitz-yuan.github.io/Myodyssey/}}}
\date{}

\begin{document}
\maketitle

\vspace{-25pt}
\begin{center}
  \includegraphics[width=\linewidth]{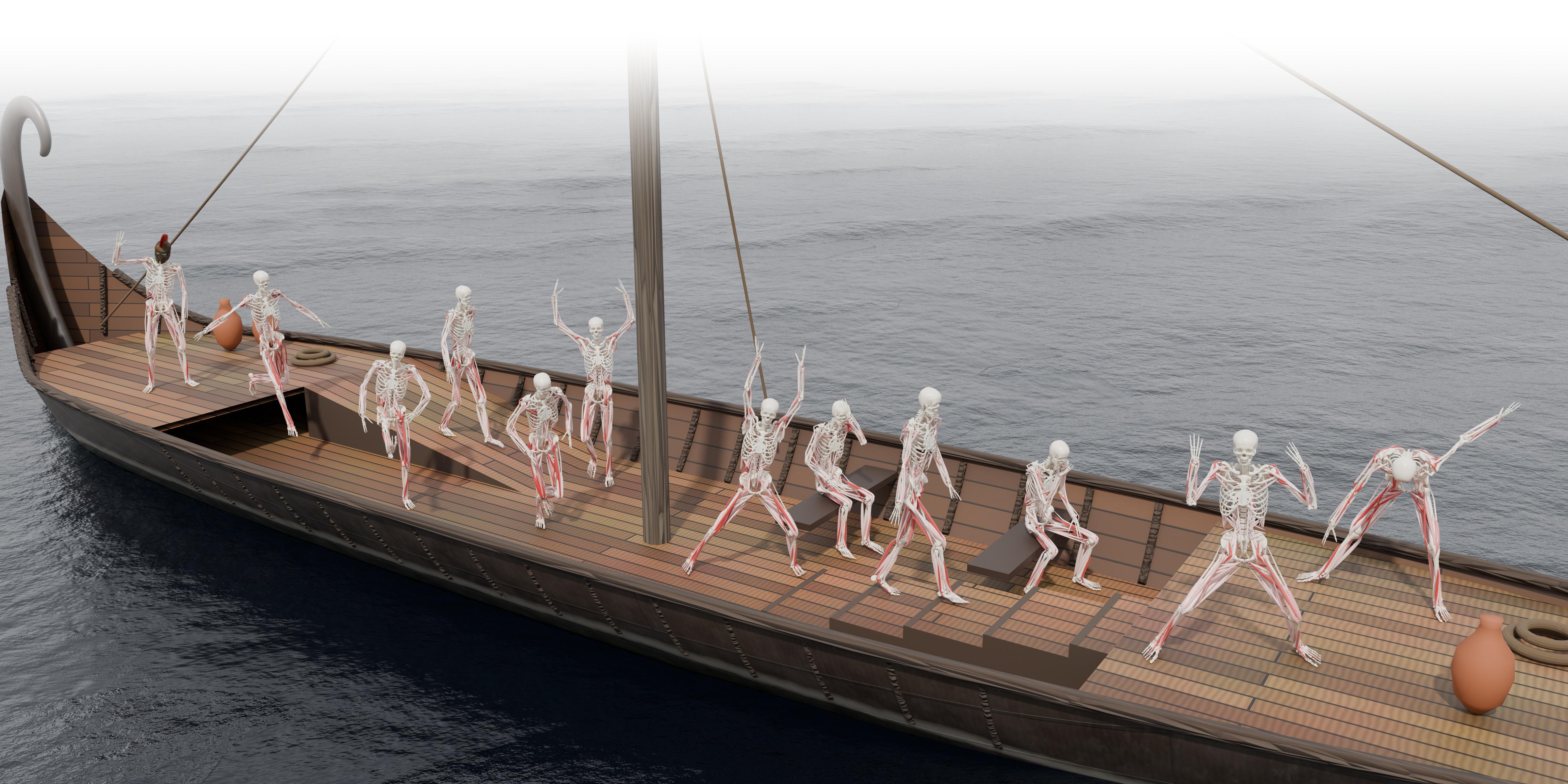}
    \captionof{figure}{\textbf{Myodyssey, a muscle-driven generalist policy trained on 182 hours of motion data, tracks a wide range of human motion on flat ground and on terrain.} On flat ground, it tracks dynamic motions such as running, dancing, jumping and kicking. On terrain, it walks down a ramp, climbs stairs, jumps off a raised deck and sits on a bench. Muscles are colored by their activation.}
    \label{fig:Myodyssey}
\end{center}

\begin{abstract}
Generalist motion-tracking policies trained on hundreds of hours of motion capture reproduce diverse human movements with humanoid robots and torque-driven characters. However, extending this capability to musculoskeletal models remains costly because of their more complex joint structures and muscle geometry, together with the need to control hundreds of redundant muscles.
We present Myodyssey, a muscle-driven motion generalist for a 354-muscle full-body model, trained through a differentiable musculoskeletal world model.
The world model computes muscle forces and joint torques analytically using a fitted surrogate of the muscle geometry and learns only the skeletal and contact dynamics from these torques, accelerating policy learning relative to an activation-input world model.
The policy likewise observes its muscles through the joint torques, making its observation invariant to activation differences in the torque-null space that produce no net torque. The policy still commands every muscle, but under the same effort penalty, this observation reduces excess muscle effort above the least-effort torque-matching reference from 232 to 38\,\%.
Trained on 175\,h of flat-ground motion, Myodyssey tracks 97.8\,\% of the motions of 8 held-out subjects and all 108 test motions from KIT, an unseen dataset, under a 0.5\,m tracking threshold.
Further training on the full 182\,h library with a local heightmap extends the policy to terrain, with 92 to 100\,\% success on stairs, ramps, seats and box steps and 81\,\% on KIT terrain, while retaining 98.2\,\% flat-ground success.
The flat-ground policy reaches about 90\,\% validation success under the 0.5\,m threshold after about 25\,min, and both training stages run to early stopping in about 11\,h on a single RTX PRO 6000 GPU.
These results show that exploiting known muscle mechanics in both dynamics learning and policy observation enables efficient training of muscle-driven generalists with broad motion coverage and lower excess muscle effort.
\end{abstract}

\section{Introduction}
\label{sec:intro}

Physics-based motion tracking is a standard approach to the control of simulated humans and humanoid robots.
A tracking policy imitates reference motions in physics simulations, so that human motions retargeted to the body become executable behaviors.
Generalist trackers trained on hundreds of hours of motion capture now serve as the low-level controllers of humanoid robots in teleoperation, motion generation and other applications~\cite{luo2026sonic,liao2026supermimic}.

A policy that tracks a captured motion with a musculoskeletal model also provides model-based estimates of muscle activity, joint torques and ground reaction forces. Conventional biomechanical analysis obtains the joint torques and muscle forces offline from measured motion and ground reaction forces with inverse dynamics and static optimization, solved separately for each recorded trial on a subject-specific model~\cite{delp2007opensim,crowninshield1981physiologically}. A tracking policy instead estimates them in simulation from the reference motion alone, for a generic body.
Muscle-driven motion tracking has scaled quickly, from about 1.8\,h of locomotion with the lower-body models of KINESIS~\cite{simos2026kinesis} to a full-body generalist on 972 KIT motions in MuscleMimic~\cite{li2026musclemimic} and to 9.4\,h of motion on flat ground, stairs, ramps, boxes and seats in TERRA~\cite{simos2026terra}.
All three are trained with proximal policy optimization (PPO)~\cite{schulman2017proximal}, which needs many simulation steps for a body with hundreds of muscles, and each uses a few hours of motion from one or a few datasets. Scaling to broader motion libraries remains a challenge.

For humanoid robots, SuperMimic~\cite{liao2026supermimic} reduced this cost by learning a single differentiable world model of the dynamics shared by all motions and training the policy with analytic gradients through short rollouts imagined in this model.
Applying this approach to a musculoskeletal model raises two questions.
First, what should the world model learn when the activation dynamics, the muscle forces and the joint torques can be computed analytically? Second, what should the policy observe about its own muscles when many different activations of hundreds of muscles produce the same torques at a few dozen joints?

This paper presents Myodyssey (\cref{fig:Myodyssey}), a framework that trains muscle-driven generalists through a differentiable musculoskeletal world model (\cref{fig:framework}), named after the long journey of one muscle-driven body through 182 hours of human motion.
The framework simulates a 354-muscle full-body model in mjlab~\cite{zakka2026mjlab} and MuJoCo Warp~\cite{mujocowarp2025} and reimplements the body and the tracking task as differentiable functions.
To address the first question, its world model combines a differentiable Hill-type muscle model, which computes the muscle forces and joint torques analytically, with a dynamics network that learns the skeletal and contact dynamics from these joint torques (\cref{sec:method}). An ablation compares this design with a world model that receives the muscle activations instead (\cref{sec:exp_obs}).

The answer to the second question affects how much muscle effort the policy uses.
In steady, well-practiced movements, humans tend to minimize their metabolic cost~\cite{donelan2001mechanical,selinger2015humans}, and effort-based criteria reproduce the main features of muscle activity in normal gait~\cite{anderson2001static,falisse2019rapid}.
Antagonist co-contraction increases in unstable or unfamiliar tasks~\cite{franklin2003adaptation} and decreases with practice, although it does not vanish~\cite{franklin2003adaptation,darainy2008muscle}.
Learned muscle-driven controllers trained without effort terms tend to co-contract strongly~\cite{schumacher2025emergence}, and current learners therefore suppress co-contraction with effort or energy terms in the reward~\cite{simos2026kinesis,feng2023musclevae,schumacher2025emergence}.

Such terms penalize co-contraction but do not remove every cause of it.
We find that a policy that observes the activations of all its muscles can learn to rely on their torque-null component, which produces no net joint torque and includes co-contraction. The tracking of such a policy degrades in a recruitment test that changes muscle-state feedback while preserving an excitation-based torque proxy (\cref{sec:exp_published}).
We refer to the dependence of a policy on how its muscles are recruited to produce a joint torque as \emph{recruitment dependence}. With the recruitment-invariant observation (RIO), our policy commands all muscles but observes their joint torques instead of their activations (\cref{sec:torque_view}).
Under the same effort penalty in the reward (\cref{sec:background}), the policy that observes every muscle uses 75 to 96\,\% more excitation than the policy with the RIO and reaches a lower tracking success under the 0.15\,m rule, 79.3 versus 85.1\,\% (\cref{sec:exp_obs}).

We train one policy on 182.4\,h of motion capture, including left--right mirrored copies, from BONES-SEED~\cite{bones2026seed}, Gait120~\cite{boo2025gait120} and a ramp-walking dataset~\cite{vielemeyer2026ramps} (\cref{sec:data}), first on flat ground and then on terrain (\cref{sec:experiments}).
The main contributions of this study are:
\begin{enumerate}
  \item \textbf{A unified retargeting pipeline that produces a large-scale motion dataset for a musculoskeletal model.} The pipeline retargets motion capture from several public datasets, together with scenes such as stairs, ramps, boxes and seats, to a 354-muscle full-body musculoskeletal model. It produces 182.4\,h of retargeted motion, to our knowledge the largest motion dataset for a musculoskeletal model.
  \item \textbf{A hybrid differentiable world model that computes the muscles analytically and learns only the skeletal and contact dynamics from the joint torques.} The differentiable muscle model computes the activation dynamics and, from the muscle lengths and moment arms of a pre-fitted geometry surrogate, the joint torques at low cost. A neural network predicts the contact-rich skeletal dynamics, for which the physics simulator (MuJoCo Warp) provides no gradients, from these joint torques. %
  \item \textbf{A recruitment-invariant observation (RIO) of the muscles.} Unlike existing work, in which the policy typically observes the state of every muscle, our policy commands every muscle but observes only the joint torques of active muscle contraction, and the passive muscle contribution follows from the observed skeletal state. The RIO reduces the reliance of the policy on the torque-null component of its activations and lowers the excess muscle effort, which includes co-contraction, from 232 to 38\,\% (\cref{sec:torque_view,sec:exp_obs}).
  \item \textbf{Myodyssey, a muscle-driven motion generalist.} Our framework trains a generalist muscle control policy that reaches high tracking success in about 11\,h of training on one GPU and has a lower excess muscle effort than the released policies of other works that we evaluated (\cref{tab:ablation,sec:exp_published}). Our policy tracks 97.8 $\pm$ 0.5\,\% of the BONES-SEED test motions of unseen subjects under the 0.5\,m rule of PHC~\cite{luo2023phc} and all 108 KIT test motions, which come from a dataset outside its training set, and after a terrain stage it also tracks motions on stairs, ramps, box steps and seats.
\end{enumerate}

\section{Related work}
\label{sec:related}

\paragraph{Motion tracking generalists.}
PHC~\cite{luo2023phc} tracks thousands of AMASS clips with a single policy.
On humanoid robots, H2O~\cite{he2024h2o}, OmniH2O~\cite{he2024omnih2o}, ExBody2~\cite{ji2026exbody2}, GMT~\cite{chen2025gmt} and BeyondMimic~\cite{liao2026beyondmimic} track retargeted human motion, and SONIC~\cite{luo2026sonic} trains a PPO tracker on 611\,h selected from about 700\,h of motion capture.
SuperMimic~\cite{liao2026supermimic} trains a tracker on 265\,h of BONES-SEED motion on one GPU through a learned world model.
All of these trackers act on joint torques or joint position targets of a rigid-body skeleton.

\paragraph{Muscle-driven motion imitation.}
MASS~\cite{lee2019scalable} imitates motions with a muscle-actuated character through a learned allocation of joint torques to muscles, and MuscleVAE~\cite{feng2023musclevae} learns a generative controller through a world model with per-muscle actions.
KINESIS~\cite{simos2026kinesis} tracks locomotion with lower-body models, MS-Human-700~\cite{zuo2024self} controls 700 muscles with hierarchical representations, and MyoAcross~\cite{wang2026myoacross} trains one tracking policy on 110 of 141 individual body models and tests it on the other 31 (\cref{tab:related}).
UniExo~\cite{yuan2026uniexo} distills four single-skill experts into one latent-conditioned policy for an 80-muscle lower-body model and fine-tunes it with reinforcement learning on transitions between the skills.
For the full body, MuscleMimic~\cite{li2026musclemimic} trains a generalist on 972 KIT motions, TERRA~\cite{simos2026terra} extends tracking to terrain with 9.4\,h of motion, and \citet{kong2026burden} train a separate 700-muscle policy for each of 24 clips. MS-Emulator~\cite{wei2026scaling} reproduces a broad repertoire of motions, including dance, cartwheels and backflips, with about 700 muscles.

\begin{table}[t]
  \centering
  \caption{\textbf{Comparison of various methods for muscle-driven motion tracking.} Unseen dataset: evaluated on a dataset that is not used in training. EMG: simulated muscle activity compared with recorded electromyography (EMG). --: not stated. BPTT: backpropagation through time~\cite{werbos1990backpropagation}. $^q$Qualitative comparison of waveforms only. $^s$Sum over 24 clips with a separate policy for each clip. $^e$Estimated from the number of KIT clips and the mean clip length of the KINESIS set (1.8\,h in 1{,}054 clips). $^m$Including left--right mirrored copies.}
  \label{tab:related}
  \small
  \resizebox{\linewidth}{!}{%
  \begin{tabular}{lcccccccc}
    \toprule
    Method & Full body & Muscles & Training motion (h) & Learner & GPU simulation & Terrain & Unseen dataset & EMG \\
    \midrule
    MuscleVAE~\cite{feng2023musclevae} & \cmark & 284 & 0.4 & Gradients in world model & \xmark & \xmark & \xmark & \xmark \\
    KINESIS~\cite{simos2026kinesis} & \xmark & 80--290 & 1.8 & PPO & \xmark & \xmark & \xmark & \cmark \\
    UniExo~\cite{yuan2026uniexo} & \xmark & 80 & 1.2 & PPO & \cmark & \xmark & \xmark & \cmark \\
    MS-Human-700~\cite{zuo2024self} & \cmark & 700 & $<$0.01 & SAC~\cite{haarnoja2018soft} & -- & \xmark & \xmark & \cmark$^q$ \\
    \citet{kong2026burden} & \cmark & 700 & 0.04$^s$ & QFlex~\cite{wei2026scalable} & -- & \xmark & \xmark & \xmark \\
    MuscleMimic~\cite{li2026musclemimic} & \cmark & 354 & 1.5 & PPO & \cmark & \xmark & \xmark & \cmark \\
    MyoAcross~\cite{wang2026myoacross} & \cmark & 416 & 0.1$^e$ & PPO & \cmark & \xmark & \xmark & \cmark \\
    TERRA~\cite{simos2026terra} & \cmark & 354 & 9.4 & PPO & \cmark & \cmark & \xmark & \cmark$^q$ \\
    muscle-supermimic~\cite{he2026musclesupermimic} & \cmark & 354--700 & 1.5 & BPTT in world model & \cmark & \xmark & \cmark & \cmark \\
    \midrule
    \rowcolor{oursrow} Myodyssey (ours) & \cmark & 354 & 182.4$^m$ & BPTT in world model & \cmark & \cmark & \cmark & \cmark \\
    \bottomrule
  \end{tabular}}
\end{table}
MuscleMimic and TERRA are trained with PPO and observe the state of every muscle.
In concurrent work, muscle-supermimic~\cite{he2026musclesupermimic} transfers SuperMimic to the MyoFullBody and MS-Human-700 models. Its policy observes the excitation and activation of every muscle, and its world model takes the activations as input, whereas our world model takes the joint torques and our policy observes the RIO. Its authors report that an effort penalty lowers the co-contraction of its policies at a cost in tracking.

\paragraph{Learning through differentiable dynamics.}
Policies can be improved by backpropagating the return through a learned model, as in stochastic value gradients~\cite{heess2015svg}, Dreamer~\cite{hafner2020dream}, SuperTrack~\cite{fussell2021supertrack} and SuperMimic~\cite{liao2026supermimic}, or through a differentiable simulator, as in SHAC~\cite{xu2022accelerated}, DiffMimic~\cite{ren2023diffmimic} and NablaMimic~\cite{lu2026nablamimic}.
MuscleVAE~\cite{feng2023musclevae} and muscle-supermimic~\cite{he2026musclesupermimic} apply a learned world model to muscle-driven bodies, with muscle target lengths and activations as their respective inputs.
Because gradients through long rollouts can explode~\cite{metz2021gradients}, SHAC and SuperMimic backpropagate through short rollouts, and our policy likewise uses 32 imagined steps (\cref{sec:learning}).
Models that combine known physics with learned components~\cite{lutter2019delan,heiden2021neuralsim} reduce what the network has to learn.
Our world model applies this idea to a musculoskeletal model. The muscle model is computed analytically, and the network learns only the contact-rich skeletal dynamics from the 31 joint torques.

\paragraph{Muscle redundancy.}
Static optimization~\cite{crowninshield1981physiologically,anderson2001static} resolves muscle redundancy by distributing the joint torques over the muscles at every instant. In human motor control, the uncontrolled manifold hypothesis~\cite{scholz1999uncontrolled} and optimal feedback control~\cite{todorov2002optimal} describe how humans mainly correct the motor errors that affect the task.
Learned controllers either structure the action space with synergies or coordination networks~\cite{berg2023sar,he2024dynsyn,lee2019scalable,park2022generative} or act on every muscle~\cite{li2026musclemimic,schumacher2023deprl,ou2026mskbench}.
\citet{jiang2019synthesis} transform a muscle-actuated optimal control problem into an equivalent problem in joint space with state-dependent torque limits. In MASS~\cite{lee2019scalable}, the control policy acts in joint space with PD targets and uses a second network to predict muscle activations that generate the resulting desired joint accelerations, while penalizing large activations to resolve the muscle redundancy.
In typical observation and learning settings for musculoskeletal control, muscle recruitment dependence can arise from muscle redundancy and we expect the RIO to reduce this dependence.

\section{Problem formulation}
\label{sec:background}
We consider a policy that tracks human motion by commanding every muscle of a full-body musculoskeletal model.
This section defines the musculoskeletal model, the muscle redundancy and the motion tracking problem.

\paragraph{Musculoskeletal model.}
We use the MyoFullBody model~\cite{li2026musclemimic} of the MyoSuite model library~\cite{caggiano2022myosuite} with the finger joints fixed at their neutral position and the 62 finger muscles removed, which leaves $m = 354$ Hill-type muscles~\cite{zajac1989muscle} with rigid tendons~\cite{millard2013flexing} acting on $n = 31$ independent joint coordinates $q$.
The policy commands an excitation $u \in [0,1]^m$, and each activation $a_i$ follows the first-order dynamics of MuJoCo~\cite{todorov2012mujoco,millard2013flexing} with the time constants of the model, $t_{\mathrm{act}} = 10$\,ms and $t_{\mathrm{deact}} = 40$\,ms for the limbs and 15 and 50\,ms for the trunk,
\begin{equation}
  \dot a_i = \frac{u_i - a_i}{t_i(a_i, u_i)}, \qquad
  t_i =
  \begin{cases}
    t_{\mathrm{act}}\,(0.5 + 1.5\,a_i) & u_i > a_i,\\
    t_{\mathrm{deact}} / (0.5 + 1.5\,a_i) & u_i \le a_i.
  \end{cases}
  \label{eq:actdyn}
\end{equation}
At the skeletal state $(q, \dot q)$, which is part of the state $s$, the muscles produce the joint torque
\begin{equation}
  \tau(s, a) = \tau_A(s, a) + \tau_P(s) = B(s)^\top a + \tau_P(s), \enspace B_{ij}(s) = -F_{0,i}\, f_L(\ell_i)\, f_V(\dot\ell_i)\, \frac{\partial \ell_i}{\partial q_j},
  \enspace \tau_{P,j}(s) = -\sum_{i=1}^{m} F_{0,i}\, f_P(\ell_i)\, \frac{\partial \ell_i}{\partial q_j}
  \label{eq:torque}
\end{equation}
with the muscle length $\ell_i$, the peak isometric force $F_{0,i}$, the force--length and force--velocity curves $f_L$ and $f_V$ of the simulator, the passive force--length curve $f_P$, and the active and passive torques $\tau_A$ and $\tau_P$.
The torque is linear in the activation, and $B(s)$ depends only on the skeletal state.

\paragraph{Muscle redundancy and effort.}
The matrix $B(s)^\top$ maps the $m = 354$ activations to $n = 31$ joint torques, and its null space $N(s) = \operatorname{null}(B(s)^\top)$ has $m - n = 323$ dimensions when $B(s)$ has rank $n$. Activations that differ by an element of $N(s)$ produce the same joint torque. Such differences include antagonist co-contraction and redistribution among redundant muscles.
At a fixed skeletal state, the instantaneous skeletal acceleration depends on muscle activation through the net joint torque. Our world model and policy observation therefore represent the muscles through their joint torques.
The tracking reward is $r = r_{\mathrm{task}} - w_e\, e(u)$, with the effort $e(u) = \frac{1}{m}\sum_{i=1}^{m} u_i^2$ and its weight $w_e = 0.1$ (\cref{app:env}).
Among the activations that produce a joint torque $\tau$, the one with the least effort is $a^{\mathrm{LE}} =\arg\min_{a \in [0,1]^m} \|a\|^2$ subject to $\tau(s, a) = \tau$. We define the torque-null component of an activation as its difference from this least-effort activation, and this difference is an element of $N(s)$.
Minimizing the squared activations is a common criterion of static optimization in biomechanics~\cite{anderson2001static}, and we use this least-effort activation $a^{\mathrm{LE}}$ as the reference when we compare the muscle effort of policies. Over the frames $t$ of a motion,
\begin{equation}
  E = \sum_t \|a_t\|^2, \qquad E_{\mathrm{LE}} = \sum_t \|a^{\mathrm{LE}}_t\|^2, \qquad \text{excess effort} = \frac{E}{E_{\mathrm{LE}}} - 1.
  \label{eq:effort}
\end{equation}
The excess effort includes co-contraction and redistribution among redundant muscles.

\paragraph{Motion tracking.}
A library $\mathcal{D}$ of reference clips gives the positions and velocities of 15 tracked bodies and the joint coordinates at every control step.
One policy $\pi_\theta$ tracks every clip of $\mathcal{D}$.
At 100\,Hz it maps an observation $o_t$ of the body and of the reference to the excitations $u_t = \pi_\theta(o_t)$ of all 354 muscles, and it maximizes the expected return $\mathbb{E}_{c \sim \mathcal{D}}\big[\sum_{t < T_c} \gamma^t r_t\big]$ with the reward $r_t$ (\cref{app:env}) and the discount factor $\gamma$.
An episode ends at the step $T_c$ at which the mean world-frame distance $d_t$ of the tracked bodies from the reference exceeds the termination distance $d_{\max} = 0.3$\,m, or at the end of the clip $c$.
We evaluate the policy by its tracking success rate on test clips outside $\mathcal{D}$ under distance thresholds of 0.5, 0.3 and 0.15\,m.

\section{Method}
\label{sec:method}

\begin{figure*}[t]
  \centering
  \includegraphics[width=\linewidth]{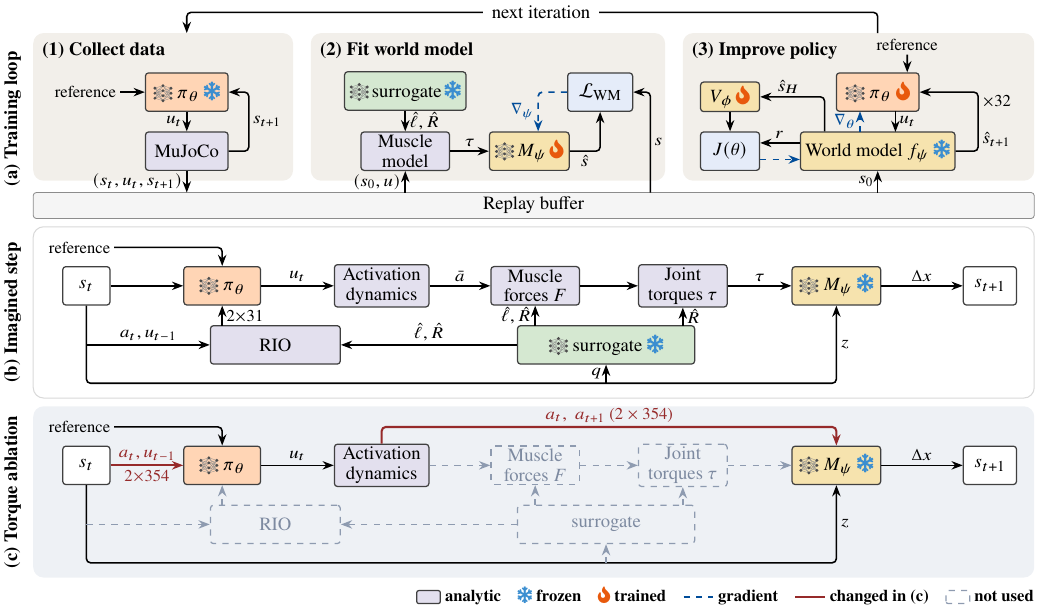}
  \caption{\textbf{Training framework and ablation overview.} (a)~Each iteration collects transitions in the simulator, fits the network $M_\psi$ of the world model $f_\psi$ on windows of 8 steps from the replay buffer, and improves the policy by backpropagating the objective $J(\theta)$ of \cref{eq:return} through 32 imagined steps of the frozen world model from buffer states $s_0$. The critic $V_\phi$, trained in the same phase, estimates the return after the last step. Hats mark predicted states, and dashed blue arrows are gradients. (b)~One imagined step of the world model $f_\psi$. The activation dynamics, the muscle forces and the joint torques are computed analytically with the frozen geometry surrogate, and $M_\psi^{\tau}$ (the superscript $\tau$ marks its joint-torque input) predicts the change $\Delta x$ of the skeletal state.
  The policy $\pi_\theta^{\tau}$ observes the body, the reference and the RIO torques (\cref{tab:obs}). (c)~The torque ablation with $(\pi_\theta^a,M_\psi^a)$. Further ablation configurations are given in \cref{sec:exp_obs}. The world model $M_\psi^a$ receives the activations $a_t$ and $a_{t+1}$ at the start and the end of the control step instead of the joint torques, and the policy $\pi_\theta^a$ observes the 354 activations $a_t$ and previous excitations $u_{t-1}$ instead of the RIO. The gray blocks are not used.}
  \label{fig:framework}
\end{figure*}

Our framework trains one tracking policy across a large motion library and follows the training loop of SuperMimic~\cite{liao2026supermimic} (\cref{fig:framework}a).
In each iteration, the current policy acts in the simulator, and the state, excitation and next state of every control step are stored in a replay buffer. The world model is then trained to predict the next state on windows of 8 steps from this buffer, and the policy is updated with the gradient of the return over rollouts that start from buffer states and are imagined in the frozen world model.
The buffer retains the data of earlier policies for training the world model, because the dynamics of the body and the scene do not depend on the policy.
The following sections describe how the simulator (\cref{sec:env}), the world model (\cref{sec:wm}), the policy and its observation (\cref{sec:torque_view}) and the policy learning (\cref{sec:learning}) are built for the musculoskeletal model.

\subsection{Simulation environment}
\label{sec:env}

The simulation environment is implemented on mjlab~\cite{zakka2026mjlab} and MuJoCo Warp~\cite{mujocowarp2025}, in which 4{,}096 parallel instances each track their own clip. The muscles are MuJoCo muscle actuators whose tendons wrap around the wrapping geometries of the model.
Following the released code of MuscleMimic~\cite{li2026musclemimic}, the simulation uses MuJoCo's Euler integrator and Newton solver with a time step of 2\,ms and 5 physics steps per control step. We make five corrections to the inertial and muscle-path data of the released model, among them one to a placeholder shoulder inertia that inflates the trunk and shoulder torques in fast arm motions, and the corrected body has a mass of 76.78\,kg (\cref{app:body}).
The motion library of 182.4\,h is stored on the GPU and sampled there (\cref{app:env}).
Each episode starts from the reference state with the muscle activations that static optimization computes for that frame, because the activation dynamics of \cref{eq:actdyn} would otherwise delay the torque needed to support a mid-motion pose.
For training on terrain, each environment can contain its own stairs, ramps, boxes and seats, and the policy then receives a local heightmap as additional input (\cref{sec:learning}).

\subsection{A world model with analytic muscles}
\label{sec:wm}

The reward, termination criterion and observation depend on the positions and velocities of the bodies and on the muscle forces.
To evaluate them on imagined states and to backpropagate through them, we reimplement the forward kinematics, the activation dynamics and the muscle forces in PyTorch.
The state $s$ consists of the root pose and velocity, the 31 independent joint coordinates and velocities and the 354 activations.
The equality constraints of the model give the dependent joints, forward kinematics gives the positions and velocities of all bodies, and the activation dynamics and muscle forces follow the muscle model of MuJoCo.
This differentiable implementation is used for both simulated and imagined states. Its muscle forces agree with those of the simulator with a relative error below $2\times10^{-6}$, and, for the same joint angles, its body positions to within $10^{-5}$\,m.

The transition from the excitation to the next state passes through the activation dynamics (i) of \cref{eq:actdyn}, the muscle geometry (ii), the muscle force (iii) and the mapping to joint torques (iv) of \cref{eq:torque}, and the response of the skeleton and its contacts (v),
\begin{equation}
  \begin{aligned}
  &\text{(i)}\;\; a^{(k+1)} = a^{(k)} + h\, \dot a\big(a^{(k)}, u\big),\; k = 0,\dots,4, \qquad
  \text{(ii)}\;\; \hat\ell = \ell_\xi(q), \;\; \hat R = R_\xi(q), \\
  &\text{(iii)}\;\; F = F_0 \odot \big[\bar a \odot f_L(\hat\ell) \odot f_V(-\hat R \dot q) + f_P(\hat\ell)\big], \qquad
  \text{(iv)}\;\; \tau = \hat R^\top F, \qquad
  \text{(v)}\;\; \Delta x = M_\psi(z, \tau),
  \end{aligned}
  \label{eq:hybridwm}
\end{equation}
where the activation is integrated over the 5 physics steps $k$ of one control step with the physics time step $h = 2$\,ms, and $\bar a$ is its mean over these steps.
The geometry surrogate with parameters $\xi$ predicts the muscle lengths $\hat\ell$ and, as a separate output, the moment-arm matrix $\hat R$ with the entries $R_{ij} = -\partial \ell_i / \partial q_j$ of \cref{eq:torque}. In (iii), $\odot$ denotes the element-wise product over the muscles.
The network $M_\psi$ predicts the change $\Delta x$ of the root and joint positions and velocities $x$ over one control step from the skeletal state $z$ in the heading frame and the joint torque $\tau$, and $x + \Delta x$ with the activation $a^{(5)}$ at the end of the step forms the next state.
Our world model evaluates (i), (iii) and (iv) analytically with the muscle parameters of MyoFullBody, (ii) with a surrogate network that is fitted before training, and (v) with the network $M_\psi$, and every operation is differentiable.

\begin{figure*}[t]
  \centering
  \includegraphics[width=\linewidth]{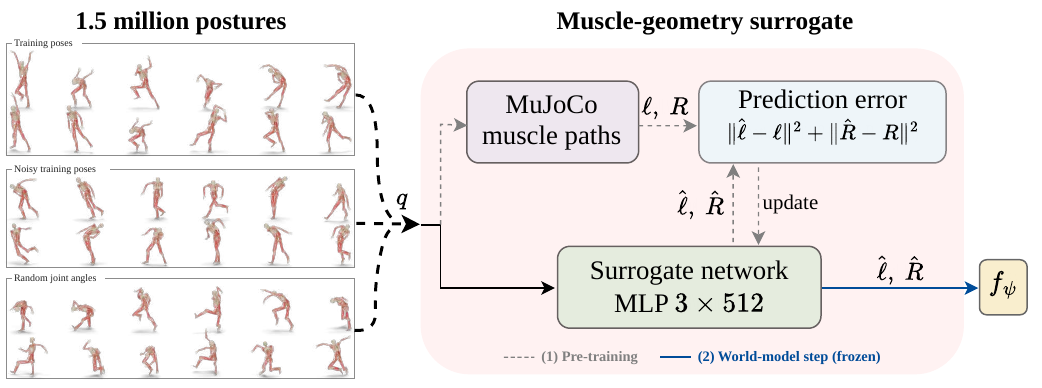}
  \caption{\textbf{Supervised training of the muscle-geometry surrogate.} The 1.5 million postures are poses from 500 training clips, with and without Gaussian noise, and random joint angles within the joint limits. $q$ denotes the joint angles of each pose. (1)~Before policy training, the surrogate network is fitted to the muscle lengths and moment arms that MuJoCo computes on these postures. (2)~During policy training, it is frozen and gives $\hat\ell$ and $\hat R$ to the world model $f_\psi$ in every step.}
  \label{fig:surrogate}
\end{figure*}

The surrogate (\cref{fig:surrogate}) replaces the tendon wrapping of MuJoCo inside the world model, which needs the muscle lengths and the moment arms, together with their gradients, in each of the 32 imagined steps of a policy update.
For a batch of 4{,}096 imagined rollouts, the combined forward and backward passes of one step take 702\,ms and 31\,GB of GPU memory with an exact differentiable implementation of the wrapping, and 8.3\,ms and 0.4\,GB with the surrogate, which is about 85 times faster.
The surrogate predicts the 354 muscle lengths and the 1{,}038 nonzero moment arms from the joint angles.
It is fitted to the simulator on 1.5 million postures (\cref{fig:surrogate}, \cref{app:env}) in about 2.5\,min on one GPU and kept fixed during training.
The median relative error of the joint torque is 0.51\,\% on the postures of the BONES-SEED test motions and 0.40\,\% on those of the KIT test motions, with 95th percentiles of 1.82 and 0.87\,\%.
On postures drawn uniformly within the joint limits, the median error is 1.6\,\%.

The skeletal and contact dynamics network $M_\psi$ is a multilayer perceptron with three hidden layers of 4{,}096 units.
It predicts the change of the root and joint positions and velocities over one control step and learns gravity, inertia, contacts, joint damping and joint limits.
Because the mapping from the 354 activations to the joint torques is computed analytically, the network does not have to learn it.
$M_\psi$ is trained on windows of 8 steps from the replay buffer and predicts each window autoregressively (\cref{app:env}).
After 32 imagined steps, its median open-loop error is 10.5\,mm for the tracked bodies and 0.88$^\circ$ for the joint angles (\cref{fig:wmacc}).

\subsection{Muscle recruitment-invariant observation (RIO) through joint torques}
\label{sec:torque_view}

The policy $\pi_\theta^{\tau}$ is a multilayer perceptron with hidden layers of 2{,}048, 2{,}048 and 1{,}024 units. It outputs the excitations of all 354 muscles.
It observes the body, the tracking error and the reference at six future steps (\cref{tab:obs}). Instead of the 354 activations and the 354 previous excitations, it observes the active joint torques $B(s)^\top a$ of the current activation and $B(s)^\top u_{t-1}$ of the previous excitation, and the passive torque follows from the observed joint angles and velocities. Because activation lags behind excitation, the torque mapped from the previous excitation provides information about the recent muscle command and potential change in torques.
We call this the recruitment-invariant observation (RIO), since activations that produce the same joint torques give the same observation. The matrix $B(s)$ is evaluated with the geometry surrogate, which keeps the observation differentiable.
The action space, the reward and the learner are the same as for a policy that observes every muscle, and the policy can still co-contract.

\begin{table}[t]
  \centering
    \caption{\textbf{Policy observation} for $\pi_\theta^{\tau}$. The root tilt, the root linear velocity and the body positions, velocities and position errors are expressed in the heading frame of the root, which is rotated from the world frame about the vertical axis by the root yaw, and the root angular velocity in the frame of the root.
    The policy $\pi_\theta^a$ of the ablation (\cref{sec:exp_obs}) replaces the two torque rows (31 dimensions each) with 354 activations $a$ and 354 previous excitations $u_{t-1}$. The steps ahead range from 0.01 to 0.4\,s with increasing intervals, dense in the near future for precise tracking and sparse further ahead for anticipating the coming motion.}
  \label{tab:obs}
  \small
  \begin{tabular}{lr}
    \toprule
    Component & Dimension \\
    \midrule
    Root height, tilt, linear and angular velocity & 13 \\
    Joint positions and velocities & 62 \\
    Positions relative to the root and velocities of 29 bodies & 174 \\
    Position error of the 15 tracked bodies & 45 \\
    Reference root height and joint velocities & 32 \\
    Reference body and joint positions at 1, 2, 5, 10, 20, 40 steps (0.01 to 0.4\,s) ahead & 456 \\
    \midrule
    Torque of the current activation, $B(s)^\top a$ & 31 \\
    Torque of the previous excitation, $B(s)^\top u_{t-1}$ & 31 \\
    \midrule
    Total & 844 \\
    \bottomrule
  \end{tabular}
\end{table}

\paragraph{Muscle recruitment dependence.}
Recruitment dependence can arise in a policy that observes its activations, although the torque-null component of the activations produces no net joint torque and is penalized by the effort term (\cref{sec:exp_published}).
In contrast, the RIO excludes the torque-null component from the observation and thereby reduces the reliance of the policy on it. This invariance is analogous to the uncontrolled manifold of human movement~\cite{scholz1999uncontrolled}, along which humans correct variations less than along the directions that affect the task.
The RIO also reduces the muscle-state input of the policy from $2 \times 354$ to $2 \times 31$ dimensions.

\paragraph{Torque-preserving muscle recruitment test.}
We use a torque-preserving muscle recruitment test to measure recruitment dependence of the control policy.
At every control step, this test projects the excitation $u_{\mathrm{pol}}$ that the policy commands onto
\begin{equation}
  u^* = \arg\min_{0 \le u \le u_{\mathrm{pol}}} \|u\|^2 \quad \text{subject to} \quad B(s)^\top u = B(s)^\top u_{\mathrm{pol}}.
  \label{eq:realloc}
\end{equation}
This projection minimizes squared excitation subject to preserving the excitation-based torque proxy $B(s)^\top u_{\mathrm{pol}}$. It can lower, but cannot increase, the excitation of any muscle. Because excitation is filtered through activation dynamics, preserving this proxy does not guarantee preservation of the actual joint torque.
We perform this least-effort projection on the excitation instead of the activation because the excitation is the immediate policy output, whereas activation evolves through dynamics included in the world model and simulator.
The projected excitation $u^*$ drives the MuJoCo simulator. The resulting least-effort activations $a(u^*)$ also enter the policy observation, either directly or through the RIO.
In a second condition, the simulator still receives $u^*$, but the muscle component of the policy observation uses shadow activations $a(u_{\mathrm{pol}})$ integrated from the unprojected commands. The skeletal observation still comes from the simulator.
We denote the tracking success without any projection by $S_a$; with the projection applied, the tracking success is $S_b$ when the policy observes $a(u_{\mathrm{pol}})$ and $S_c$ when it observes $a(u^*)$.
The recruitment-dependence index $\mathrm{RDI} = (S_b - S_c)/S_b$ is the relative drop in tracking success when the policy observes $a(u^*)$ instead of $a(u_{\mathrm{pol}})$. Comparing $S_c$ with $S_b$, rather than with the unprojected baseline $S_a$, holds the projection procedure fixed while changing the source of muscle-state feedback, isolating the effect of policy observation and its dependency on muscle recruitment.

\subsection{Policy learning}
\label{sec:learning}

The policy maximizes the return of $H = 32$ steps imagined in the world model $f_\psi$ of \cref{eq:hybridwm} from states of the replay buffer,
\begin{equation}
  J(\theta) = \sum_{t=0}^{H-1} \gamma^t c_t\, r_t + \gamma^H c_H\, V_\phi(o_H), \qquad
  s_{t+1} = f_\psi(s_t, u_t), \quad c_{t+1} = c_t\,(1 - \delta_{t+1}).
  \label{eq:return}
\end{equation}
Here $\gamma$ is the discount factor, $V_\phi$ is the critic, fitted to TD($\lambda$) targets, and $c_t$ is the weight with which the episode is still running at step $t$, with $c_0 = 1$.
The termination is smoothed as $\delta_t = \operatorname{sig}\big((d_t - d_{\max})/(0.02\,\text{m})\big)$ with the logistic function $\operatorname{sig}$, so that the return is differentiable.
The reward $r_t$ is the tracking reward of \cref{sec:background}, which combines exponential tracking terms of the body positions and velocities and of the joint positions and velocities with penalties on the mean squared excitation, the change of the policy output and the joint acceleration (\cref{app:env}).
In a muscle-driven body, a few imagined trajectories can have sensitivities that grow along the 32 steps and dominate the gradient of backpropagation through time (BPTT).
To prevent this, the gradient through the imagined state of each sample is limited in norm to 10 times the median norm over the batch.

Data are collected for 2 control steps per iteration, with clips sampled in proportion to their duration.
Three quarters of the environments add Gaussian noise of standard deviation 0.1 to the policy output and clip the result.
Because clipped noise raises the excitation of the muscles that the deterministic policy keeps inactive, these noisy environments do not fully cover the states that the deterministic policy visits. A fixed quarter of the environments therefore execute the deterministic policy, which is also the policy that is evaluated.
The learning rate of $2\times10^{-4}$ is lowered once to $6\times10^{-5}$ when the success on a validation set of BONES-SEED subjects stops improving.
Training ends when this success stops improving again, and the checkpoint is selected on the same set by a rule fixed before the final runs (\cref{app:stages}).

The terrain stage starts from this checkpoint and adds a local heightmap of $11\times11$ points with a spacing of 0.1\,m to the inputs of the policy, the critic and the world model.
The new input branches are initialized to zero, so that the policy at the start of the stage equals the flat-ground policy.
In this stage, the policy observes a perturbed reference in 50\,\% of the episodes, and the reward and the critic use the clean reference (\cref{app:env}).
This stage trains on the whole training set (\cref{sec:data}), with 1{,}024 of the 4{,}096 environments fixed to the terrain clips, and uses the same learning-rate schedule, plateau rule and checkpoint selection, applied to the mean of the success on a flat-ground and a terrain validation set.

To speed up training, the world-model, policy and critic updates are each captured as one CUDA graph, and the imagined step is compiled with \texttt{torch.compile}.
The updates run on a separate CUDA stream while the next collection proceeds, as in muscle-supermimic~\cite{he2026musclesupermimic}.
With these changes, a training iteration takes 1.31\,s on one GPU (\cref{app:compute}).

\section{Data and evaluation}
\label{sec:data}

\paragraph{A unified retargeting pipeline.}
The same retargeting code is used for every source and takes the marker set, the support geometry and the subject calibration as input.
Gait120 and the ramp dataset provide optical markers.
For BONES-SEED and KIT, whose joints differ from those of MyoFullBody, we place virtual markers on the released skeleton or body mesh, so that the inverse kinematics matches body positions instead of copying joint angles.
Each subject is calibrated on standing frames with the feet flat.
The geometry of floors, stairs, ramps, boxes and seats is taken from the measurements of the dataset or fitted to the motion, and the joint coordinates are solved by inverse kinematics on the GPU~\cite{yuan2026mocaptomyo}.
If a physical check fails, for example for a foot that penetrates the ground, floats or slides, a window of frames is solved again with contact constraints, and the new solution is kept if it adds no marker-fit failure.
Frames that still fail are cut, acrobatic motions are removed, and a quality filter with thresholds fixed before training is applied to every clip (\cref{app:quality}).

\paragraph{Training and test sets.}
The training set contains 182.4\,h of motion in 115{,}789 clips from 599 subjects (\cref{tab:data}).
BONES-SEED provides left--right mirrored copies of its motions, and we add mirrored copies to the training sets of the other sources.
Without the mirrors, the training set contains 91.1\,h.
The flat-ground stage is trained on the flat-ground motions of BONES-SEED~\cite{bones2026seed}, from which the quality filter removes 6.1\,\% of the training hours.
The terrain stage adds 7.7\,h of motion on stairs, ramps, boxes and seats.
BONES-SEED contributes motions onto and off seats and motions onto, off and over boxes, and 89\,\% of the box clips are steps onto and off a box of 50\,cm.
Gait120~\cite{boo2025gait120} and \citet{vielemeyer2026ramps} contribute ramp walking, and Gait120 also contributes stair walking.
Each training source is split by subject into training, validation and test sets, and a subject and its mirrored motions are assigned to the same set.
The seat sets are an exception. Because the BONES-SEED test subjects have no seat motions, the seat validation and test sets consist of seat motions of eight training subjects that are not used for training.
Two further test sets come from KIT, a dataset outside the training sources.
The KIT test set consists of the 108 KIT test motions of MuscleMimic~\cite{li2026musclemimic,mandery2015kit,mahmood2019amass}.
The KIT terrain set consists of the KIT motions on stairs and ramps with the scenes of TERRA~\cite{simos2026terra}, and every clip that passes the quality filter is used for testing.

\begin{table}[t]
  \centering
    \caption{\textbf{Motion data} in hours after the quality filter, with the number of subjects in parentheses. Mirrored copies make up about half of the hours of the training sets and of all BONES-SEED sets, and the other validation and test sets contain no mirrored copies. The box and seat subjects also appear in the flat-ground sets of BONES-SEED, and the total counts every subject once within each column. The box validation and test subjects belong to the flat-ground validation and test sets. The KIT test motions are the test split of KINESIS and MuscleMimic, and the KIT terrain set contains the KIT stair and ramp trials of AMASS, 14.8\,min from 7 subjects before the quality filter.}
  \label{tab:data}
  \small
  \begin{tabular}{lccc}
    \toprule
    Source & Training & Validation & Test \\
    \midrule
    BONES-SEED, flat ground & 174.6 (491) & 4.7 (10) & 2.0 (8) \\
    BONES-SEED, boxes & 2.94 (345) & 0.09 (10) & 0.12 (5) \\
    BONES-SEED, seats & 2.64 (63) & 0.24 (4) & 0.10 (4) \\
    Gait120, stairs and ramps & 1.72 (98) & 0.07 (9) & 0.10 (10) \\
    Vielemeyer, ramps & 0.43 (10) & 0.01 (1) & 0.05 (2) \\
    \midrule
    Total & 182.4 (599) & 5.1 (24) & 2.4 (24) \\
    \midrule
    KIT, 108 test motions & -- & -- & 0.16 (21) \\
    KIT, terrain & -- & -- & 0.16 (7) \\
    \bottomrule
  \end{tabular}
\end{table}

\paragraph{Metrics.}
Each test clip is played once from its first frame with the deterministic policy and with the termination of the environment switched off.
The success rule of PHC and of most tracking generalists~\cite{luo2023phc,he2024omnih2o,simos2026kinesis} counts a clip as failed at the first frame at which the mean world-frame distance $d_t$ of the tracked bodies from the reference exceeds 0.5\,m.
We also report this rule with the stricter thresholds of 0.3 and 0.15\,m, where 0.3\,m equals the termination distance $d_{\max}$ of training.
Under the root-relative rule of MuscleMimic~\cite{li2026musclemimic}, denoted root-rel.\ 0.5\,m, a clip fails when the mean root-relative deviation of the tracked bodies or the deviation of the pelvis exceeds 0.5\,m.
A clip counts as a fall when the height of the pelvis above the ground stays more than 0.3\,m below that of the reference for 1\,s, and the no-fall rate is the percentage of clips without a fall.
Following PHC~\cite{luo2023phc}, we report the global and root-relative position errors $E_\text{g-mpjpe}$ and $E_\text{mpjpe}$ and the velocity and acceleration errors $E_\text{vel}$ and $E_\text{acc}$.
The reported excitation is the mean over the 354 muscles.
Results are given as the mean and standard deviation over the training seeds.

\section{Experiments}
\label{sec:experiments}

We first test whether released muscle-driven policies and our policies depend on the torque-null component of their muscle activation (\cref{sec:exp_published}). We then evaluate the tracking accuracy on test subjects and across datasets (\cref{sec:exp_main}), the effect of the RIO and of the torque input of the world model on tracking and effort (\cref{sec:exp_obs}), the agreement of the excitation, the ground reaction forces and the joint torques in level walking with human data (\cref{sec:exp_emg}), the extension of tracking to stairs, ramps, boxes and seats by the terrain stage (\cref{sec:exp_terrain}), and the robustness of tracking to pushes that are not applied during training (\cref{sec:exp_push}).

The main configuration is trained with three random seeds on one NVIDIA RTX PRO 6000 Blackwell Max-Q GPU per run, and its checkpoint is selected on the validation set by the rule of \cref{app:stages}.
The ablation varies the input of the world model and the observation of the policy, each of which uses either the joint torques (superscript $\tau$) or the muscle activations (superscript $a$).
Myodyssey is the policy $\pi_\theta^\tau$ trained with the world model $M_\psi^\tau$, and the ablation trains $\pi_\theta^\tau$ with $M_\psi^a$, $\pi_\theta^a$ with $M_\psi^\tau$ and $\pi_\theta^a$ with $M_\psi^a$.
All configurations share the data, the flat-ground stage and the evaluation, and are compared with Myodyssey on the same test clips.

\begin{figure}[t]
  \centering
  \includegraphics[width=\linewidth]{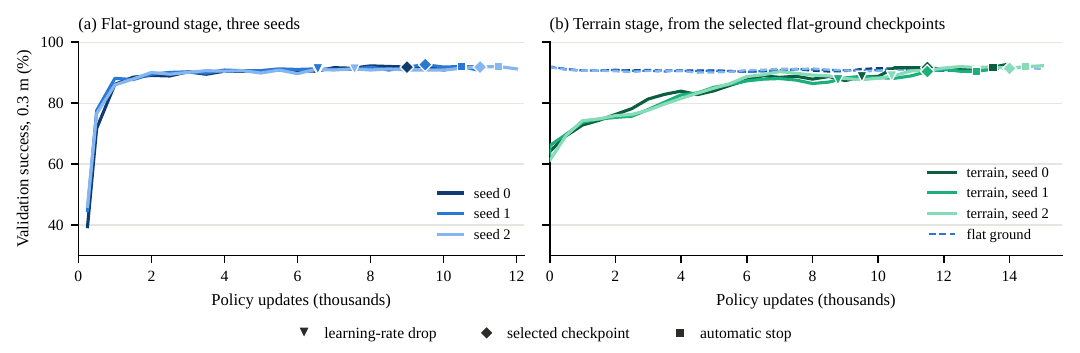}
    \caption{\textbf{Validation success during training in two stages.} Validation success of the deterministic policy under the 0.3\,m rule for three seeds, with markers for the learning-rate decay, the selected checkpoint and early stopping. (a)~Flat-ground stage. All three seeds exceed 85\,\% after 1{,}000 policy updates, about 25\,min of training. (b)~Terrain stage, continued from the selected flat-ground checkpoint of each seed. Solid lines show the terrain validation set and dashed lines the flat-ground validation set, smoothed over three evaluations.}
  \label{fig:stages}
\end{figure}

In both stages (\cref{fig:stages}), the learning-rate decay, early stopping and checkpoint selection are triggered by rules fixed before the final runs.
Excluding the evaluations of the validation set, the flat-ground stage reaches its selected checkpoint after about 3.6\,h of training on one GPU and stops after 4.2\,h, and the terrain stage after 5.8 and 6.8\,h, about 11\,h in total until early stopping (\cref{app:compute} and \cref{tab:train_cost}). After 1{,}000 policy updates of the flat-ground stage, about 25\,min of training, the validation success already reaches about 90\,\% under the 0.5\,m rule and exceeds 85\,\% under the 0.3\,m rule.
Under the 0.3\,m rule, the terrain stage raises the success on the terrain validation set from 62.8 to 91.8\,\% on average over the three seeds, while the flat-ground validation success stays around 91\,\%.

\subsection{Joint-torque observation reduces reliance on redundant activation}
\label{sec:exp_published}

The recruitment test projects the muscle recruitment commanded by the policy ($u_{\mathrm{pol}}$) onto the least-effort recruitment ($u^*$), as defined in \cref{eq:realloc}.
Learned activations generally contain a torque-null component. A drop in success when the policy observes activations integrated from the projected excitations rather than shadow activations indicates sensitivity to the change of the torque-null component in the muscle-state feedback.
We applied the test to the released policies of four published full-body models, each with its own environment, motions and success rule, and to our final flat-ground policies $\pi_\theta^\tau$ and $\pi_\theta^a$, both trained with $M_\psi^\tau$ (\cref{tab:published}).

Under the projection, the success of MuscleMimic falls from 88.0\,\% with the shadow activations $a(u_{\mathrm{pol}})$ to 9.0\,\% with the least-effort activations $a(u^*)$, and 12 of the 14 MSK-Bench policies reach an RDI of at least 0.97.
These policies were trained with PPO, SAC and DEP-RL, so the dependence also arises in learners that do not backpropagate through the observation.
TERRA and MS-Human-700 maintain their success under the projection that recruits no new muscles, so observing the activations does not always lead to this dependence. However, our evaluation of their released environments reveals excess effort (\cref{eq:effort}) of 106 and 1{,}243\,\%, respectively.

At a fixed skeletal state, the RIO maps activations that differ only within the torque-null space to the same observation, excluding direct dependence on that component.
For the Myodyssey default configuration $(\pi_\theta^\tau, M_\psi^\tau)$, the success under the three conditions is nearly identical (98.9, 98.8 and 98.8\,\%).
Its RDI is 0.000 with a 95\,\% interval of $[-0.003, 0.003]$ over three seeds and 300 test clips per seed, and the indices of the individual seeds lie between $-0.003$ and $0.007$ under both the 0.5 and the 0.3\,m rule.
The policy $\pi_\theta^a$ of the ablation $(\pi_\theta^a, M_\psi^\tau)$ observes the activations that the projection changes and behaves in the opposite way. Under the projection, its success falls from 97.6\,\% with the shadow activations to 10.2\,\% with the least-effort activations, and to 6.2\,\% under the 0.3\,m rule. Its RDI is 0.895 with an interval of $[0.868, 0.921]$ over three seeds, and the indices of the individual seeds lie between 0.787 and 0.955. In a normal rollout, this policy encounters least-effort activations only at the first frame (\cref{sec:env}), after which the activations follow its own excitations, whereas the recruitment test projects the excitations at every control step.

\begin{table}[t]
  \centering
  \caption{\textbf{Muscle recruitment test.} Success $S$ in \% without the least-effort projection ($S_a$) and with the least-effort projection ($u^*$) while the policy observes the shadow activations $a(u_{\mathrm{pol}})$ ($S_b$) or the least-effort activations $a(u^*)$ ($S_c$). The recruitment-dependence index (RDI) is given with a 95\,\% bootstrap interval.
  Myodyssey $(\pi_\theta^\tau, M_\psi^\tau)$ and the ablation $(\pi_\theta^a, M_\psi^\tau)$ produce the final flat-ground policies for the 354-muscle MyoFullBody, evaluated on 300 BONES-SEED test motions per seed under the 0.5\,m rule. For other models, we test the released policies in their respective code environments, using the motions and success criteria provided by the original implementations. The muscle state is the component of the observation that describes the muscles, and every policy also observes the body and the reference (\cref{tab:obs} for Myodyssey).}
  \label{tab:published}
  \small
  \setlength{\tabcolsep}{4pt}
  \resizebox{\linewidth}{!}{%
  \begin{tabular}{lllcccc}
    \toprule
    Model & Learner & Observed muscle state & $S_a$ & $S_b$ & $S_c$ & RDI \\
    \midrule
    MuscleMimic~\cite{li2026musclemimic} & PPO & length, velocity, force, excitation, activation & 88.3 & 88.0 & 9.0 & 0.90 [0.85, 0.94] \\
    TERRA~\cite{simos2026terra} & PPO & excitation, activation & 95.7 & 94.5 & 93.5 & 0.01 [$-$0.01, 0.03] \\
    MS-Human-700~\cite{zuo2024self} & SAC & length, velocity, force, activation & 95 & 95 & 93 & 0.02 [$-$0.04, 0.08] \\
    MSK-Bench~\cite{ou2026mskbench}, 14 policies & SAC, DEP-RL & length, velocity, force, activation & 28--100 & 12--98 & 0--20 & $\geq 0.97$ in 12 of 14 \\
    \midrule
    $(\pi_\theta^a, M_\psi^\tau)$, 3 seeds & world model & activation, previous excitation & 97.2 & 97.6 & 10.2 & 0.895 [0.868, 0.921] \\
    \rowcolor{oursrow} Myodyssey $(\pi_\theta^\tau, M_\psi^\tau)$, 3 seeds & world model & RIO & 98.9 & 98.8 & 98.8 & 0.000 [$-$0.003, 0.003] \\
    \bottomrule
  \end{tabular}}
\end{table}

\subsection{One policy for everyday motion}
\label{sec:exp_main}

\begin{table}[t]
  \centering
    \caption{\textbf{Tracking on flat ground} with the deterministic policy, without action noise, mean $\pm$ standard deviation over three seeds. Success rates in \%. The errors are averaged over all frames of all clips, with positions in mm, velocities in m/s and accelerations in m/s$^2$ from finite differences at 100\,Hz. Root-rel.\ 0.5\,m is the root-relative rule of MuscleMimic (\cref{sec:data}).}
  \label{tab:main}
  \small
  \setlength{\tabcolsep}{3pt}
  \resizebox{\linewidth}{!}{%
  \begin{tabular}{lccccccccc}
    \toprule
     & \multicolumn{5}{c}{Success} & \multicolumn{4}{c}{Error} \\
    \cmidrule(lr){2-6}\cmidrule(lr){7-10}
    Test set & 0.5\,m & 0.3\,m & 0.15\,m & Root-rel.\ 0.5\,m & No-fall & $E_\text{g-mpjpe}$ & $E_\text{mpjpe}$ & $E_\text{vel}$ & $E_\text{acc}$ \\
    \midrule
    BONES-SEED, test subjects (1{,}574) & 97.8 $\pm$ 0.5 & 96.5 $\pm$ 0.4 & 85.1 $\pm$ 1.2 & 97.8 $\pm$ 0.5 & 98.1 $\pm$ 0.6 & 58.0 $\pm$ 1.9 & 36.6 $\pm$ 0.7 & 0.233 $\pm$ 0.003 & 6.54 $\pm$ 0.02 \\
    KIT, test motions (108) & 100.0 $\pm$ 0.0 & 100.0 $\pm$ 0.0 & 89.2 $\pm$ 4.2 & 100.0 $\pm$ 0.0 & 100.0 $\pm$ 0.0 & 46.5 $\pm$ 6.2 & 19.4 $\pm$ 0.6 & 0.144 $\pm$ 0.007 & 4.22 $\pm$ 0.05 \\
    BONES-SEED, training subset (512) & 96.9 $\pm$ 0.5 & 96.5 $\pm$ 0.4 & 88.9 $\pm$ 1.7 & 97.0 $\pm$ 0.6 & 97.8 $\pm$ 0.6 & 58.4 $\pm$ 5.4 & 33.9 $\pm$ 1.1 & 0.170 $\pm$ 0.004 & 4.80 $\pm$ 0.04 \\
    \bottomrule
  \end{tabular}}
\end{table}

\Cref{tab:main} reports the flat-ground policy on two test sets and a training subset.
The 8 BONES-SEED test subjects contribute 1{,}574 clips that are used neither for training nor for the checkpoint selection, the 108 KIT test motions come from a dataset outside the training set, and a fixed subset of 512 training clips measures how well the policy fits its training data.
Under the 0.5\,m rule, the policy tracks 97.8\,\% of the BONES-SEED test clips and all KIT test motions.
Under the 0.15\,m rule, it tracks 85.1 and 89.2\,\% of them, with root-relative errors $E_\text{mpjpe}$ of 36.6 and 19.4\,mm.
Under the 0.5 and 0.3\,m rules, the success on the test subjects is as high as on the training subset, which indicates that the policy generalizes to the motions of new subjects. Only under the 0.15\,m rule is it lower (85.1 versus 88.9\,\%). With Gaussian action noise of standard deviation 0.1, the success under the 0.5\,m rule is 96.8\,\% on the BONES-SEED test subjects and 100\,\% on the KIT test motions (\cref{app:noise}).

\begin{figure*}[t]
  \centering
  \includegraphics[width=\linewidth]{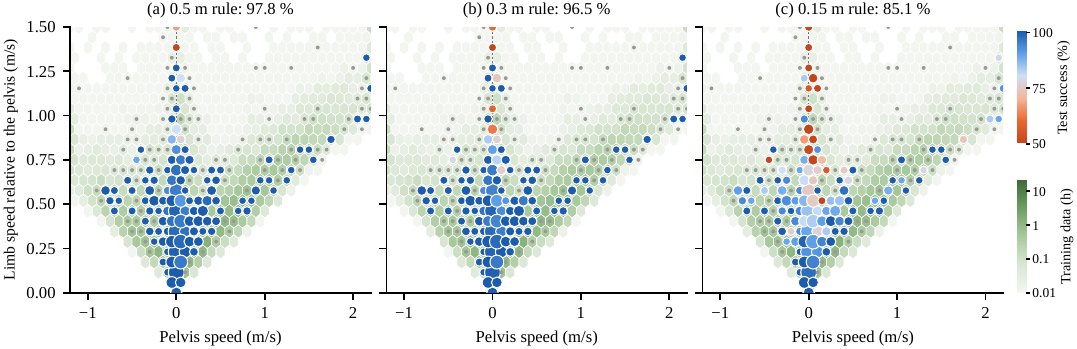}
  \caption{\textbf{Test motions in motion space.} Each BONES-SEED test motion is placed by the mean speed of the pelvis along the facing direction of the body, negative for backward travel, and the mean speed of the tracked bodies relative to the pelvis in its reference. The green hexagons give the hours of training data in each cell. The circles give the test motions in a cell, with a size that increases with their number and a color for their success under the 0.5, 0.3 and 0.15\,m rules in (a) to (c), averaged over the motions and the three seeds. Cells with fewer than three test motions are gray dots.}
  \label{fig:motionspace}
\end{figure*}

The failures depend on the kind of motion. On the BONES-SEED test subjects, the success under the 0.5\,m rule is 96.2\,\% for dancing, and the dances with the fastest limb movement account for most of the failures, whereas the other categories reach 99.0 to 100\,\% (\cref{app:detail}).
The test subjects perform no motions close to the ground such as crawling, so the test does not cover them.
\Cref{fig:motionspace} places every test motion by the forward speed of its pelvis and the speed of its limbs on a density map of the training data.
Under the 0.5 and 0.3\,m rules, nearly every motion is tracked, including the motions with backward travel.
Under the strict 0.15\,m rule, the policy still tracks nearly every motion in regions where the training data are dense, and its failures concentrate on fast limb motions with little travel of the pelvis, such as dance moves, where the training data are sparser.

KINESIS~\cite{simos2026kinesis} and MuscleMimic~\cite{li2026musclemimic} report tracking on the 108 KIT test motions of the KINESIS split after training on its KIT training motions, whereas our policy is trained without KIT motions (\cref{tab:musclemimic}).
KINESIS controls lower-body models with 80 to 290 muscles, and MuscleMimic controls the MyoFullBody model.
Each work retargets and processes the motions in its own way, so the table serves as a reference and not as a controlled comparison.
Under the success rule of each work, Myodyssey tracks all test motions, and its global position error of 46.5\,mm lies between the 42.8 and 48.8\,mm that KINESIS reports for the four configurations of its lower-body models.
Under the error definitions of MuscleMimic, our policy has lower joint-angle, joint-velocity, root-position and site errors than MuscleMimic reports, and a higher root-yaw error.
We also evaluate the flat-ground policy on all 1{,}080 KIT motions that MuscleMimic uses, its 972 training and 108 test motions. The policy tracks 99.5 $\pm$ 0.1\,\% of them under the 0.5\,m rule, 99.1 $\pm$ 0.2\,\% under the 0.3\,m rule and 84.3 $\pm$ 2.4\,\% under the 0.15\,m rule.

\begin{table*}[t]
  \centering
    \caption{\textbf{KIT test motions.} Reported values of KINESIS (its best model Legs+Back with 290 muscles, mean $\pm$ standard error over three evaluation runs) and MuscleMimic, and values of Myodyssey with the deterministic policy as the mean $\pm$ standard deviation over three seeds. All values refer to the 108 KIT test motions of the KINESIS split, and each work processes the motions in its own way. --: not reported. Frame coverage is the percentage of frames before the first failure under the rule of each work. Global position is the mean distance of the tracked bodies from the reference. The other errors follow the definitions in the code of MuscleMimic, evaluated at the 17 bodies that carry its mimic sites. The best value of each column is in bold.}
  \label{tab:musclemimic}
  \small
  \newcommand{\hd}[1]{\begin{tabular}[b]{@{}c@{}}#1\end{tabular}}
  \setlength{\tabcolsep}{4pt}
  \resizebox{\textwidth}{!}{%
  \begin{tabular}{lccccccccccc}
    \toprule
    & & \multicolumn{3}{c}{Success (\%)} & \multicolumn{7}{c}{Error} \\
    \cmidrule(lr){3-5}\cmidrule(lr){6-12}
    Policy & \hd{KIT in\\training} & 0.5\,m & \hd{root-rel.\\0.5\,m} & \hd{frame\\coverage} & \hd{global\\position (mm)} & \hd{joint\\angle ($^\circ$)} & \hd{joint\\velocity ($^\circ$/s)} & \hd{root\\position (cm)} & \hd{root\\yaw ($^\circ$)} & \hd{relative\\site (cm)} & \hd{absolute\\site (cm)} \\
    \midrule
    KINESIS (Legs+Back), reported & yes & 96.91 $\pm$ 0.50 & -- & 99.43 $\pm$ 0.03 & \textbf{42.84} $\pm$ 0.29 & -- & -- & -- & -- & -- & -- \\
    MuscleMimic, reported & yes & -- & 92.62 $\pm$ 0.01 & 96.04 $\pm$ 0.45 & 128 & 6.63 $\pm$ 0.01 & 26.92 $\pm$ 0.03 & 7.11 $\pm$ 0.15 & \textbf{1.77} $\pm$ 0.03 & 2.28 $\pm$ 0.02 & 12.85 $\pm$ 0.24 \\
    \rowcolor{oursrow} Myodyssey (ours) & no & \textbf{100.0} $\pm$ 0.0 & \textbf{100.0} $\pm$ 0.0 & \textbf{100.0} $\pm$ 0.0 & 46.5 $\pm$ 6.2 & \textbf{3.91} $\pm$ 0.29 & \textbf{18.5} $\pm$ 0.1 & \textbf{2.35} $\pm$ 0.38 & 3.39 $\pm$ 0.02 & \textbf{1.24} $\pm$ 0.04 & \textbf{4.33} $\pm$ 0.64 \\
    \bottomrule
  \end{tabular}}
\end{table*}

\subsection{Ablation: joint-torque observation reduces muscle effort and accelerates learning}
\label{sec:exp_obs}
\label{sec:exp_wm}

\begin{table}[t]
  \centering
  \caption{\textbf{Ablations and baseline comparisons.} Our policies are evaluated with deterministic actions. Success in \% under the three rules and mean excitation. The excess effort (\cref{eq:effort}) is computed on 300 BONES-SEED test clips and on the 108 KIT test motions. Each of our configurations is the mean of three seeds. MuscleMimic is the released checkpoint, trained on KIT motions and evaluated on its own retargeted motions with sampled actions, as in its evaluation. We train muscle-supermimic using its released code and the fast recipe in its paper~\cite{he2026musclesupermimic} for 20\,min on the KIT training motions, with three seeds, and evaluate it with deterministic actions.}
  \label{tab:ablation}
  \small
  \resizebox{\linewidth}{!}{%
  \begin{tabular}{llccccc}
    \toprule
    Test set & Configuration & 0.5\,m & 0.3\,m & 0.15\,m & Excitation & Excess effort (\%) \\
    \midrule
    \rowcolor{oursrow} BONES-SEED, test subjects & Myodyssey $(\pi_\theta^\tau,M_\psi^\tau)$ & 97.8 & 96.5 & 85.1 & 0.208 & 38 \\
     & $(\pi_\theta^a,M_\psi^\tau)$ & 97.1 & 95.1 & 79.3 & 0.363 & 232 \\
     & $(\pi_\theta^\tau,M_\psi^a)$ & 97.1 & 95.5 & 81.8 & 0.235 & 71 \\
     & $(\pi_\theta^a,M_\psi^a)$ & 97.3 & 95.0 & 78.0 & 0.358 & 240 \\
    \midrule
    \rowcolor{oursrow} KIT, 108 test motions & Myodyssey $(\pi_\theta^\tau,M_\psi^\tau)$ & 100.0 & 100.0 & 89.2 & 0.173 & 26 \\
     & $(\pi_\theta^a,M_\psi^\tau)$ & 99.4 & 99.1 & 87.0 & 0.340 & 237 \\
     & $(\pi_\theta^\tau,M_\psi^a)$ & 100.0 & 99.7 & 85.8 & 0.195 & 56 \\
     & $(\pi_\theta^a,M_\psi^a)$ & 99.1 & 98.8 & 82.1 & 0.333 & 247 \\
     \cmidrule{2-7}
     & MuscleMimic~\cite{li2026musclemimic} & 88.9 & 66.4 & 23.8 & 0.421 & 440 \\
     & muscle-supermimic~\cite{he2026musclesupermimic} & 82.7 & 55.9 & 17.3 & 0.408 & 343 \\
    \bottomrule
  \end{tabular}}
\end{table}

The policy $\pi_\theta^a$ observes the 354 activations and the 354 previous excitations in place of the RIO (\cref{fig:framework}c). The world model $M_\psi^a$ replaces the joint torques in its input by the 354 activations at the start and the end of each control step, as in muscle-supermimic~\cite{he2026musclesupermimic}. All configurations use the same remaining settings, data and evaluation as Myodyssey and are trained with three seeds (\cref{tab:ablation}).
To measure how much the activation exceeds the least-effort activation, we compute $a^{\mathrm{LE}}$ by static optimization with the muscle model of the simulator at every frame of 300 BONES-SEED test clips and of the 108 KIT test motions, and evaluate \cref{eq:effort} over the frames of each clip before its first 0.3\,m failure.
The excess effort is averaged over the clips and seeds, with the clips of each KIT motion averaged first.
All four configurations reach 97.1 to 97.8\,\% on the BONES-SEED test subjects under the 0.5\,m rule, and the two choices mainly change the muscle effort and the success under the 0.15\,m rule.
The policy observation accounts for most of the difference in effort. With either world model, observing the activations raises the excess effort to 232 or 240\,\%, whereas changing the world-model input has a smaller effect (from 38 to 71\,\% with the RIO).

\begin{figure}[t]
  \centering
  \includegraphics[width=\linewidth]{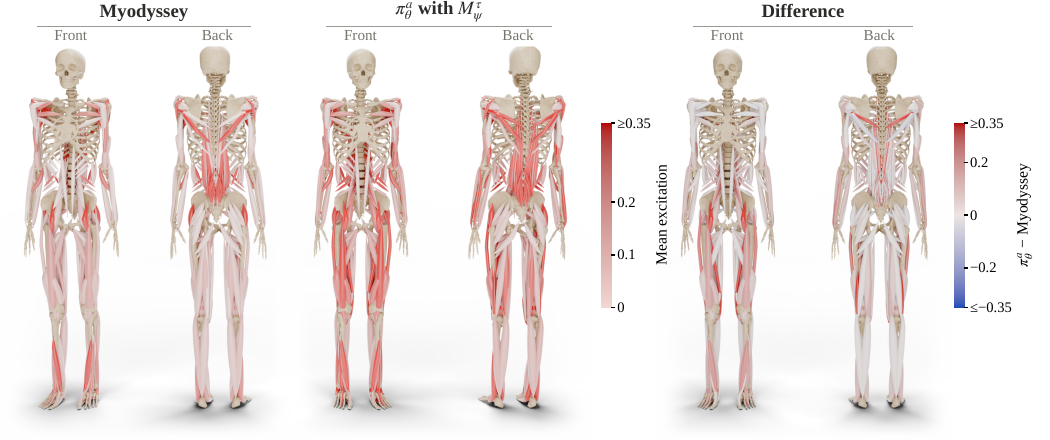}
    \caption{\textbf{Mean excitation of every muscle} over 300 BONES-SEED test motions (by clip name 57\,\% dance, 35\,\% gait and 8\,\% jumps and other motions) for Myodyssey (left) and $\pi_\theta^a$ with $M_\psi^\tau$ (middle), one training run each with the same seed, with left and right muscles averaged, and their difference (right). Both track nearly the same percentage of these motions (98.7 and 97.7\,\% under the 0.5\,m rule). The muscles that $\pi_\theta^a$ activates more strongly than Myodyssey are mainly the back extensors (quadratus lumborum, longissimus, multifidus), the psoas, the short head of the biceps femoris and the latissimus dorsi.
    }
  \label{fig:musclemap}
\end{figure}

\paragraph{Policy observation ($\pi_\theta^a$ with $M_\psi^\tau$).}
With the same effort term, the policy $\pi_\theta^a$ tracks the test motions almost as well as Myodyssey but has a lower success under the 0.15\,m rule on the BONES-SEED test subjects (79.3 versus 85.1\,\%).
Its mean excitation is 0.363 versus 0.208 on the BONES-SEED test subjects and 0.340 versus 0.173 on the KIT test motions.
Its excess effort is 232\,\%, compared with 38\,\% for the RIO. It also learns more slowly and first reaches 90\,\% validation success under the 0.3\,m rule after $5{,}700 \pm 1{,}600$ policy updates, compared with $2{,}500 \pm 500$ with the RIO.
The recruitment test shows that this policy depends on the torque-null component of its activations (\cref{sec:exp_published}), which the RIO excludes from the observation (\cref{sec:torque_view}).
The additional activation of $\pi_\theta^a$ does not make it more robust to pushes. Half of its motions fall at $125 \pm 3$\,N\,s, versus $126 \pm 7$\,N\,s with the RIO (\cref{sec:exp_push}).
Most of the additional excitation is in the back and the hip (\cref{fig:musclemap}). Of this excitation, 50\,\% is in the extensors of the back and 20\,\% in the muscles of the thigh and the psoas.

\paragraph{World-model input ($\pi_\theta^\tau$ with $M_\psi^a$).}
With $M_\psi^a$, the final policy tracks the test motions nearly as well as Myodyssey but has a lower success under the 0.15\,m rule (81.8 versus 85.1\,\%). It uses 13\,\% more excitation on the BONES-SEED test subjects, and its excess effort is 71\,\%, compared with 38\,\% for the joint-torque input.
The policy also learns more slowly with $M_\psi^a$ and first reaches 90\,\% validation success under the 0.3\,m rule after $4{,}300 \pm 1{,}300$ policy updates, compared with $2{,}500 \pm 500$ with the joint-torque input, although both reach about 92\,\% at their selected checkpoints.

\subsection{Muscle activity and kinetics in walking}
\label{sec:exp_emg}
\begin{figure}[!htb]
  \centering
  \includegraphics[width=\linewidth]{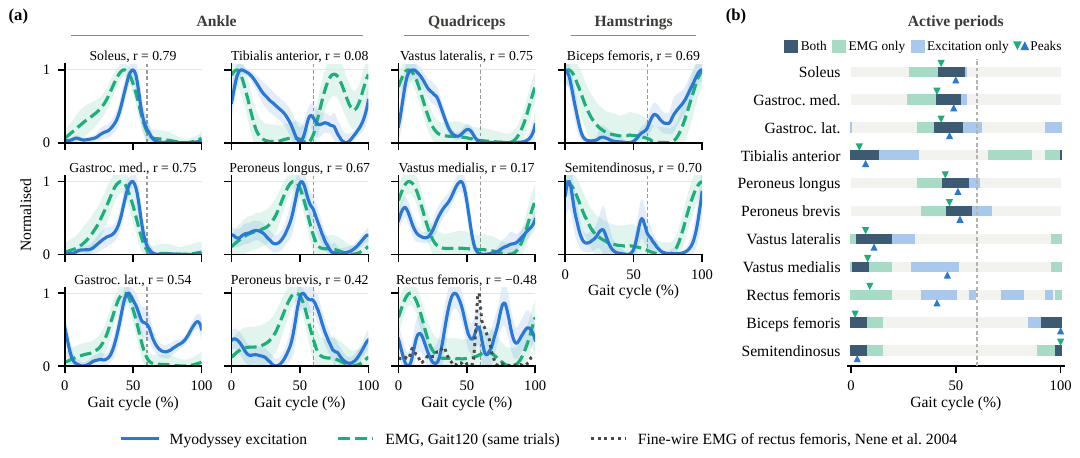}
    \caption{\textbf{Muscle activity of the right leg in level walking} on 469 Gait120 trials, which are not used for training, with the excitation averaged over three seeds, the muscles grouped into ankle, quadriceps and hamstring muscles, and toe-off as the dashed line. (a)~Mean cycle over 101 subjects with $\pm 1$ standard deviation, scaled to its range for display, with the correlation $r$ of the subject-averaged curves in the title and the fine-wire EMG recording of \citet{nene2004rectus} dotted. (b)~Active phases, where the scaled mean cycle is above 0.5, and the peaks of the mean cycles as triangles.}
  \label{fig:emg}
\end{figure}

We compare the excitation of the right-leg muscles of the flat-ground policies with the electromyography (EMG) of Gait120~\cite{boo2025gait120}, whose level-walking trials are not used for training (\cref{fig:emg}).
The flat-ground policies track all 538 level-walking clips of 116 subjects under the 0.5\,m rule from their first frame.
The right heel strikes on the force plates of the dataset define the same gait cycles for the EMG and for the simulation. We exclude the first 0.5\,s and the last 0.4\,s of each trial from both signals to remove the effects of the episode start and the end of the reference, which leaves 805 cycles from 469 trials of 101 subjects.
We compare the excitation and not the activation, since surface EMG reflects the neural drive~\cite{buchanan2004neuromusculoskeletal}, and divide the mean cycle of each subject by its peak, as is common for EMG~\cite{burden2010normalize}.

The correlation between the simulated and the measured mean cycle of each subject is 0.37 on average over the 11 muscles and the three seeds, and between 0.34 and 0.42 for individual seeds.
For comparison, the correlation between one subject and the mean of the other subjects is 0.88.
The value of 0.37 lies within the range of 0.2 to 0.6 that MuscleMimic reports for walking~\cite{li2026musclemimic}, although the definitions differ in detail.
MSK-Bench~\cite{ou2026mskbench} shifts each simulated cycle circularly to maximize the correlation. With this shift limited to 20\,\% of the gait cycle, the mean correlation rises to 0.64.
This value measures the shape of the cycle apart from its timing, so a substantial part of the disagreement lies in the timing of the muscle activity.
The ankle plantar flexors, for example, become active 8 to 14\,\% of the cycle later than in the EMG (\cref{fig:emg}b), whereas the vastus lateralis and the hamstrings are active in early stance as in the EMG.
The rectus femoris and the vastus medialis show an additional burst in mid stance at 41 to 46\,\% of the cycle, and the tibialis anterior stays active until 32\,\% of the cycle and lacks the burst of the EMG in swing.
For the rectus femoris, the surface recording itself is uncertain. Its surface EMG in walking is contaminated by cross-talk from the vastus intermedius at initial contact, whereas fine-wire recordings within the muscle show little activity there and a burst around toe-off~\cite{nene2004rectus}.
Like the fine-wire recording, the simulated excitation of the rectus femoris stays below half of its range in early stance and has a burst near toe-off.
Agreement with EMG is lower when the policy observes activations. With $\pi_\theta^a$ and $M_\psi^\tau$, the correlation on the same trials is $0.27 \pm 0.05$ over three seeds, compared with $0.37 \pm 0.05$ for Myodyssey.

\begin{figure}[!htb]
  \centering
  \includegraphics[width=\linewidth]{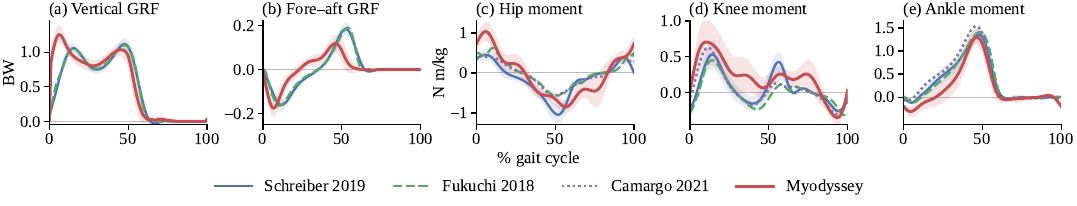}
  \caption{\textbf{Ground reaction forces and joint torques in level walking.} Right leg over a gait cycle from heel strike, for Myodyssey on the Gait120 level-walking trials and for three human datasets~\cite{schreiber2019multimodal,fukuchi2018public,camargo2021comprehensive}. Myodyssey shows the joint torque of its muscles, averaged over three seeds. Bands give $\pm 1$ standard deviation across subjects, and torques are positive for extension and plantarflexion.}
  \label{fig:kinetics}
\end{figure}

On the Gait120 level-walking trials, the ground reaction forces and the joint torques of the muscles also agree in shape with human data (\cref{fig:kinetics}, \cref{app:kinetics}).
Compared with the speed-matched data of \citet{schreiber2019multimodal}, the correlation is $0.89 \pm 0.01$ for the vertical force and $0.86 \pm 0.00$, $0.76 \pm 0.06$ and $0.95 \pm 0.03$ for the hip, knee and ankle torques.
The fore--aft force leads the human force by about 6\,\% of the cycle, and its correlation of $0.72 \pm 0.03$ rises to $0.93 \pm 0.01$ when it is shifted. The first peak of the vertical force is higher and earlier than in humans (1.25 versus 1.06 body weights).

\subsection{The same policy extends to terrain}
\label{sec:exp_terrain}
\begin{figure}[t]
  \centering
  \includegraphics[width=\linewidth]{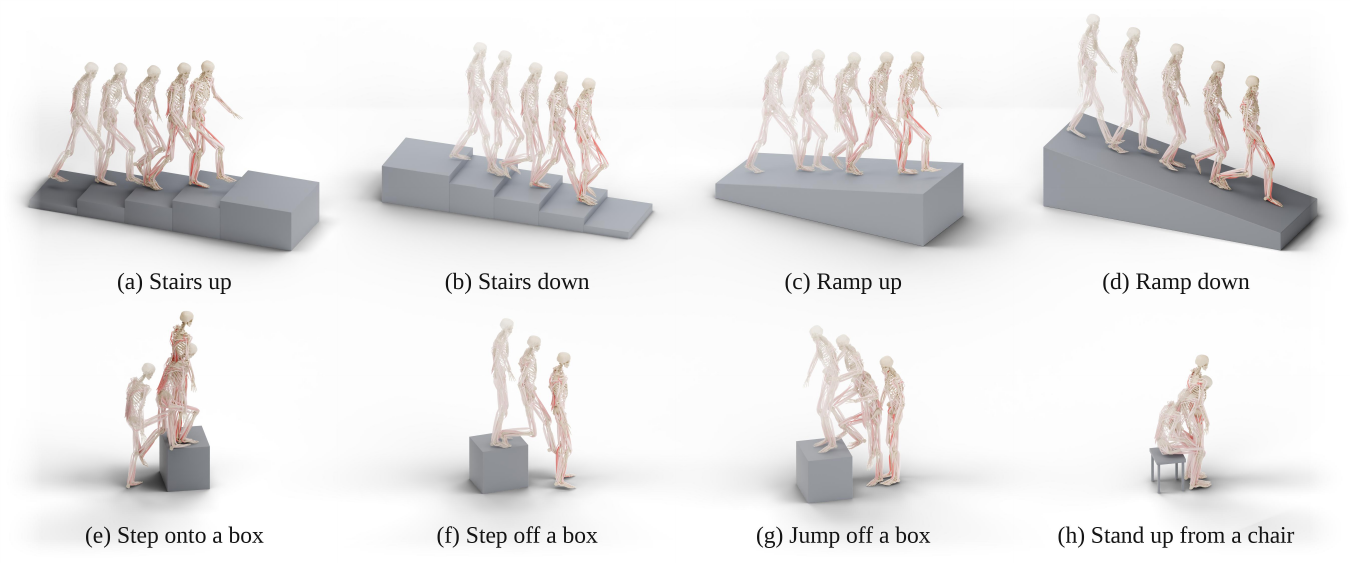}
    \caption{\textbf{Motions on terrain.} Test motions from Gait120 (a, b), Vielemeyer (c, d) and BONES-SEED (e--h), tracked by the Myodyssey policy after the terrain stage in simulation, with the muscles colored by their activation.}
  \label{fig:terrain}
\end{figure}

\Cref{tab:terrain} compares the policy after the terrain stage with the flat-ground policy from which the stage starts, as the mean over three seeds, and \cref{fig:terrain} shows examples of the tracked motions.
The test sets are the test motions of the terrain sources (\cref{sec:data}) and the KIT terrain motions, which come from a dataset outside the training set.
Under the 0.5\,m rule, the policy tracks 92.0\,\% of the stair and ramp motions of the Gait120 test subjects, all motions of the Vielemeyer test subjects and 98.7\,\% of the seat test motions, compared with 87.8, 100 and 48.4\,\% for the flat-ground policy.
On the KIT terrain motions, the success rises from 55.1 to 81.0\,\%.
The flat-ground policy, which has no heightmap, already tracks most stair and ramp motions under the 0.5\,m rule (98\,\% for stair descent, 63\,\% for stair ascent and 99\,\% for Gait120 ramps), and on these motions the terrain stage mainly improves the accuracy. Under the 0.15\,m rule, the success rises from 57.3 to 87.0\,\% on Gait120 and from 47.4 to 93.7\,\% on the Vielemeyer ramps.
On the boxes, the policy tracks 94.7\,\% of the 44 motions that step onto and off a box of 50\,cm, but only 35.1\,\% of the 97 jumps onto, off and over boxes, and such jumps make up 8\,\% of the box clips in the training set.
After the terrain stage, the policy still tracks 98.2\,\% of the BONES-SEED test motions and 99.7\,\% of the KIT test motions under the 0.5\,m rule, but its success under the 0.15\,m rule on the BONES-SEED test motions decreases from 85.1 to 80.7\,\%.
The mean excitation on the BONES-SEED test subjects is 0.209 after the stage and 0.208 before it.
With action noise, the success under the 0.5\,m rule falls most on the boxes, to 45.6\,\%, and on the KIT terrain motions, to 76.9\,\%, and stays at 90.8\,\% or higher on the other terrain sets (\cref{app:noise}).

\begin{table}[t]
  \centering
  \caption{\textbf{Tracking on terrain} with the deterministic policy, without action noise, mean $\pm$ standard deviation over three seeds. Success in \% under the 0.5, 0.3 and 0.15\,m rules. MPJPE in mm in the world frame is averaged over the frames before the first 0.3\,m failure and therefore differs from $E_\text{g-mpjpe}$ in \cref{tab:main}. The flat-ground policy is the checkpoint from which the terrain stage starts.}
  \label{tab:terrain}
  \small
  \resizebox{\linewidth}{!}{%
  \begin{tabular}{lccccccccc}
    \toprule
     & & \multicolumn{3}{c}{Myodyssey (terrain stage)} & \multicolumn{3}{c}{Flat-ground policy} & \multicolumn{2}{c}{MPJPE} \\
    \cmidrule(lr){3-5}\cmidrule(lr){6-8}\cmidrule(lr){9-10}
    Test set & Clips & 0.5\,m & 0.3\,m & 0.15\,m & 0.5\,m & 0.3\,m & 0.15\,m & Terrain stage & Flat-ground policy \\
    \midrule
    Gait120, stairs and ramps, test subjects & 167 & 92.0 $\pm$ 0.3 & 90.4 $\pm$ 0.0 & 87.0 $\pm$ 0.9 & 87.8 $\pm$ 2.5 & 85.4 $\pm$ 2.4 & 57.3 $\pm$ 10.0 & 59.8 $\pm$ 2.8 & 85.8 $\pm$ 7.6 \\
    Vielemeyer, ramps, test subjects & 95 & 100.0 $\pm$ 0.0 & 100.0 $\pm$ 0.0 & 93.7 $\pm$ 9.1 & 100.0 $\pm$ 0.0 & 100.0 $\pm$ 0.0 & 47.4 $\pm$ 23.2 & 54.1 $\pm$ 5.7 & 81.2 $\pm$ 16.4 \\
    BONES-SEED seats, test motions & 51 & 98.7 $\pm$ 1.1 & 98.7 $\pm$ 1.1 & 94.1 $\pm$ 2.0 & 48.4 $\pm$ 11.8 & 41.2 $\pm$ 10.9 & 24.2 $\pm$ 15.8 & 37.9 $\pm$ 2.3 & 98.3 $\pm$ 3.8 \\
    BONES-SEED boxes, test subjects & 141 & 53.7 $\pm$ 2.2 & 49.9 $\pm$ 2.3 & 37.6 $\pm$ 0.0 & 26.5 $\pm$ 0.4 & 26.0 $\pm$ 0.4 & 20.8 $\pm$ 0.4 & 60.0 $\pm$ 0.7 & 73.8 $\pm$ 0.8 \\
    KIT terrain, not used for training & 156 & 81.0 $\pm$ 1.3 & 77.1 $\pm$ 3.2 & 64.7 $\pm$ 2.3 & 55.1 $\pm$ 3.4 & 49.6 $\pm$ 3.2 & 43.4 $\pm$ 0.7 & 56.1 $\pm$ 2.0 & 70.9 $\pm$ 1.7 \\
    \midrule
    BONES-SEED flat ground, test subjects & 1{,}574 & 98.2 $\pm$ 0.3 & 95.3 $\pm$ 0.3 & 80.7 $\pm$ 0.5 & 97.8 $\pm$ 0.5 & 96.5 $\pm$ 0.4 & 85.1 $\pm$ 1.2 & 50.7 $\pm$ 1.2 & 47.1 $\pm$ 1.5 \\
    KIT, 108 test motions & 108 & 99.7 $\pm$ 0.5 & 98.5 $\pm$ 0.5 & 92.3 $\pm$ 2.8 & 100.0 $\pm$ 0.0 & 100.0 $\pm$ 0.0 & 89.2 $\pm$ 4.2 & 41.0 $\pm$ 3.7 & 45.2 $\pm$ 5.4 \\
    \bottomrule
  \end{tabular}}
\end{table}

\subsection{Recovery from unseen pushes}
\label{sec:exp_push}

\begin{figure}[!htb]
  \centering
  \includegraphics[width=\linewidth]{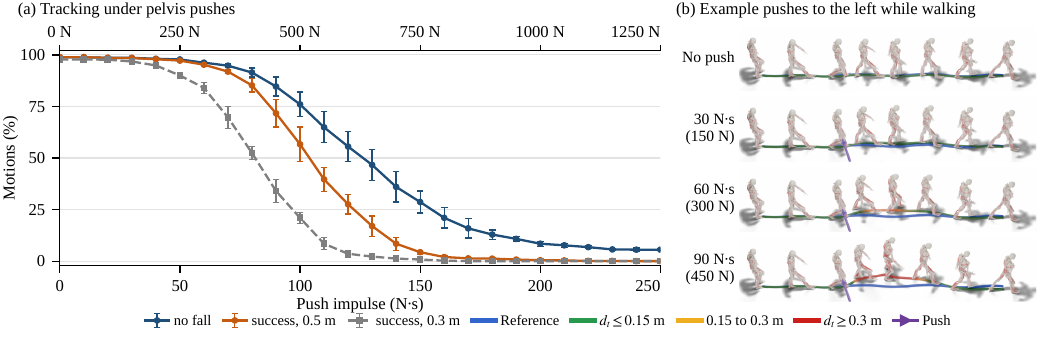}
  \caption{\textbf{Robustness to pushes} on the pelvis, which are never applied in training. (a)~Success and no-fall rate on 300 BONES-SEED test motions, each pushed once for 0.2\,s at a random time and in a random direction, against the impulse and the force, with the mean and standard deviation over three seeds. (b)~A walking motion pushed to the left at the same moment with increasing impulse, with the paths of the pelvis in the reference and in the simulation, colored by the tracking error $d_t$.}
  \label{fig:push}
\end{figure}

We push each of the 300 BONES-SEED test motions of \cref{tab:ablation} once at the pelvis with a horizontal force for 0.2\,s at a random time and in a random direction (\cref{fig:push}, \cref{app:push}). Pushes are never applied during training, so the test measures the response to external forces that the policy has not experienced.
Up to 40\,N\,s (200\,N), the tracking stays almost unchanged, and half of the motions fall at $126 \pm 7$\,N\,s (630\,N), which corresponds to 1.64\,N\,s per kilogram of body mass. Over the four directions of the push, this impulse ranges from 116\,N\,s for pushes from the front to 134\,N\,s for pushes to the left.
In the example of \cref{fig:push}b, the body returns to the reference path after the push of 90\,N\,s, and the tracking error falls back below 0.15\,m within 1.7\,s.

\section{Discussion and limitations}
\label{sec:limitations}
The recruitment test and the ablation show that the observation of the policy provides a way to reduce activation beyond what the joint torques require.
The RIO does not restrict the action space, since the policy still commands every muscle and can co-contract, but it removes the torque-null component from what the policy observes. The RIO also reduces the muscle-related input from 708 to 62 dimensions, although our ablation does not separate the effect of the invariance from that of the smaller input.
The RIO and the recruitment test can be applied with any learner. We applied the test to released PPO, SAC and DEP-RL policies (\cref{tab:published}) but did not train a model-free policy with the RIO.
Training model-free policies with the RIO would test whether its benefits extend beyond world-model-based learning.
Why the dependence arises in MuscleMimic and the MSK-Bench policies but not in TERRA and MS-Human-700 remains open. These two policies nevertheless have a high excess effort (\cref{sec:exp_published}), so the dependence is not the only source of excess effort in learned controllers. Uncertainty is another possible source, since co-contraction can minimize the effort in a system with noise~\cite{koelewijn2022antagonistic}.

The muscle model simplifies the human musculotendon unit to support efficient simulation and learning. The muscles have rigid tendons. When the tendon is shorter than the optimal fiber length, rigid tendons produce forces similar to those of elastic tendons and can be simulated 2 to 54 times faster with explicit integration, but their force error grows with tendon length~\cite{millard2013flexing}. The muscles also have no pennation angle, and their parameters and moment arms are taken from one generic model and do not capture the variation between individuals~\cite{li2026musclemimic}.
Extending the differentiable muscle model to include tendon compliance and pennation, together with subject-specific calibration, could improve its biomechanical fidelity.

The world model requires a differentiable implementation of the muscle model and a geometry surrogate fitted to the body, and adapting to a new body or muscle model requires updating these components.
The policy currently controls one body with one set of proportions and fixed fingers.
Conditioning the policy and world model on body parameters and training across varied bodies could extend motion tracking to bodies of different sizes.
We did not train a model-free learner on the same data, since published PPO trackers take about 60 to 240 GPU hours on 0.1 to 9.4\,h of motion (\cref{tab:train_cost}), so the training time of Myodyssey is not compared with such a learner under the same conditions. 

Myodyssey reaches about 90\,\% validation success under the 0.5\,m rule after about 25\,min while training on 175\,h of motion without KIT data (\cref{sec:experiments}), showing that broad motion coverage can be learned early in training.
Compared with our three-seed reproduction of muscle-supermimic using its released fast recipe on 1.5\,h of KIT motion, Myodyssey achieves higher KIT test success under all three distance thresholds and lower excess effort (\cref{tab:ablation}).
In extended 1\,h runs of muscle-supermimic, the test success does not improve, and the return plateaus after around 20\,min.
These results compare the evaluated configurations; differences in training data, retargeting and control rate prevent attributing the performance gap to a single design choice.
The fast recipe of muscle-supermimic uses Beta exploration noise and 50\,Hz control, whereas Myodyssey uses Gaussian noise and 100\,Hz control.
Evaluating these choices within our framework could identify further efficiency gains.

The policy tracks given reference motions and could be coupled with a motion generator for tasks without a predefined reference.
In walking, the first peak of the vertical ground reaction force is higher and earlier than in humans.
Incorporating kinetic measurements into training or model calibration could help address this discrepancy.
Jumps onto, off and over boxes remain difficult, and all terrain is static.
Future work will add more dynamic motions and more varied terrain to the training data and extend the policy to bodies of different sizes.

\section{Conclusion}
\label{sec:conclusion}

We presented Myodyssey, a muscle-driven motion generalist for a 354-muscle full-body model, trained by backpropagation through a differentiable musculoskeletal world model.
By computing muscle forces and joint torques explicitly, the world model focuses learning on skeletal and contact dynamics, while the policy observes muscle state through joint torques and retains direct control of every muscle.
The flat-ground policy reaches about 90\,\% validation success under the 0.5\,m rule after about 25\,min.
The full two-stage training uses 182\,h of motion and runs to early stopping in about 11\,h on a single RTX PRO 6000 GPU, yielding a single policy that generalizes to held-out subjects and an unseen dataset and tracks motions on flat ground and terrain.
Beyond tracking performance, the results show that the policy observation shapes muscle recruitment.
Under the same effort penalty, policies with similar tracking success can use substantially different muscle effort; observing joint torques reduces excess effort from 232 to 38\,\% in our ablation and lowers reliance on redundant activation.
This finding makes the representation of muscle state an important design choice alongside the reward.
Its benefits have been demonstrated on one body model with world-model-based learning, and remain to be established with model-free learners and separated from the effect of the smaller observation dimension.
Myodyssey provides a basis for studying muscle recruitment across diverse movements, with broader biomechanical use requiring validation against human measurements and extension to varied bodies.

\section*{Data and model availability}
Motion Data by Bones Studio is available at \url{https://bones.studio/}. The KIT motions are used through AMASS~\cite{mahmood2019amass} under its non-commercial research license, and Gait120~\cite{boo2025gait120} and the ramp dataset of \citet{vielemeyer2026ramps} are used under CC BY 4.0. The terrain of the KIT motions comes from TERRA~\cite{simos2026terra}. The walking data of \citet{schreiber2019multimodal}, \citet{fukuchi2018public} and \citet{camargo2021comprehensive} are used under CC BY 4.0. No motion data are redistributed.
The MyoFullBody model is distributed under the Apache License 2.0 and builds on OpenSim~\cite{delp2007opensim} models~\cite{rajagopal2016full,christophy2012musculoskeletal}, among them the MoBL-ARMS upper-extremity model~\cite{saul2015benchmarking}, whose use is restricted to non-commercial purposes.

\bibliographystyle{plainnat}
\bibliography{refs}

\clearpage
\appendix
\section*{Appendix}

\section{Muscle recruitment test details}
\label{app:realloc}
The least-effort projection solves the quadratic program of \cref{eq:realloc} at every control step, for our policies with a dual semi-smooth Newton method of 40 iterations, with $B(s)$ of \cref{eq:torque} computed from the muscle lengths, muscle velocities and moment arms of the simulator. The solution $u^*$ is sent to the simulator in place of $u_{\mathrm{pol}}$, and the simulator computes the activations ($a(u^*)$) and the body motion that determine the tracking success. We use this solver for the recruitment test and for $E_{\mathrm{LE}}$ (\cref{eq:effort}), and not in training.
On the other hand, we obtain $a(u_{\mathrm{pol}})$ by integrating $u_{\mathrm{pol}}$ with the same activation dynamics as MuJoCo.
The main text reports the projection with $0 \le u \le u_{\mathrm{pol}}$, which lowers each muscle and recruits no new muscle.
When the constraint is relaxed to $0 \le u \le 1$ (\cref{tab:realloc}), the solution can also recruit muscles that are inactive under $u_{\mathrm{pol}}$. Under this variant, TERRA keeps most of its success with the shadow activations but loses part of it with the activations from the projected excitations (RDI 0.26), so the absence of a dependence for TERRA in \cref{sec:exp_published} holds only for the variant that recruits no new muscles. For MS-Human-700, the relaxed projection itself lowers the success to 17\,\%, so its index of 1.00 under this variant rests on few successful episodes.
We test every released policy in its respective code environment using its own motions, action mode and success rule, and the episodes of the three conditions are paired.

\begin{table}[h]
  \centering
  \caption{\textbf{Protocols and results of the recruitment test} for both upper bounds of the projection ($u \le u_{\mathrm{pol}}$ and $u \le 1$), in \%. $S_a$ is the success that we measured without the projection, with the released checkpoint for each published model evaluated in its own code environment, and the value in parentheses is the success that each work reports in its publication. $S_b$ and $S_c$ are the success with the projection while the policy observes $a(u_{\mathrm{pol}})$ and $a(u^*)$, respectively. The RDI is given with its 95\,\% bootstrap interval. --: not reported.}
  \label{tab:realloc}
  \small
  \setlength{\tabcolsep}{4pt}
  \resizebox{\linewidth}{!}{%
  \begin{tabular}{llllccccccc}
    \toprule
     & & & & & \multicolumn{3}{c}{$u \le u_{\mathrm{pol}}$} & \multicolumn{3}{c}{$u \le 1$} \\
    \cmidrule(lr){6-8}\cmidrule(lr){9-11}
    Model & Episodes & Actions & Success rule & $S_a$ (reported) & $S_b$ & $S_c$ & RDI & $S_b$ & $S_c$ & RDI \\
    \midrule
    MuscleMimic & 108 KIT clips $\times$ 3 & sampled & root-rel.\ 0.5\,m & 88.3 (92.6) & 88.0 & 9.0 & 0.90 [0.85, 0.94] & 45.4 & 2.8 & 0.94 [0.88, 0.99] \\
    TERRA & 135 KIT clips $\times$ 5 & sampled & 0.25\,m & 95.7 (91.7) & 94.5 & 93.5 & 0.01 [$-$0.01, 0.03] & 92.0 & 67.7 & 0.26 [0.20, 0.33] \\
    MS-Human-700 & 100 seeds & sampled & 3 gait cycles & 95 (--) & 95 & 93 & 0.02 [$-$0.04, 0.08] & 17 & 0 & 1.00 [1.00, 1.00] \\
    MSK-Bench, 14 policies & 50 seeds per policy & deterministic & MSK-Bench task & 28--100 & 12--98 & 0--20 & $\geq 0.97$ in 12 of 14 & 8--100 & 0--22 & $\geq 0.98$ in 13 of 14 \\
    \midrule
    $\pi_\theta^a$ with $M_\psi^\tau$, 3 seeds & 300 BONES-SEED clips $\times$ 3 & deterministic & 0.5\,m & 97.2 & 97.6 & 10.2 & 0.895 [0.868, 0.921] & 97.4 & 6.0 & 0.938 [0.916, 0.959] \\
    Myodyssey, 3 seeds & 300 BONES-SEED clips $\times$ 3 & deterministic & 0.5\,m & 98.9 & 98.8 & 98.8 & 0.000 [$-$0.003, 0.003] & 98.8 & 98.8 & 0.000 [$-$0.005, 0.004] \\
    \bottomrule
  \end{tabular}}
\end{table}
\section{Environment, world model and reward}
\label{app:env}

\paragraph{Motion library.}
Each environment tracks its own clip from its own start time, and a motion library on the GPU provides the reference of every environment at every step and at six future steps.
To hold the 31 million frames of the flat-ground training set on the GPU, the motions are stored at 50\,Hz and interpolated to the control rate, and the joint coordinates on the device are kept in half precision.
Start poses are read from a single-precision table, so that every start satisfies the joint equality constraints.
The start activations are computed before training by static optimization and stored with the motions.
Episodes end when $d_t$ exceeds $d_{\max}$, at the end of the clip or after 40\,s.

\paragraph{Terrain.}
Each environment holds up to 13 boxes, and these boxes represent the stairs, ramps, boxes and seats of its clip.
A ray-cast sensor measures the heightmap of $11\times11$ points with a spacing of 0.1\,m around the root, aligned with its heading.
The terrain validation set contains 348 clips of the four terrain sources, from their validation subjects and, for the seats, from held-out seat motions of training subjects (\cref{sec:data}).

\paragraph{Reference perturbation.}
The perturbed reference is a perturbed pose passed through the forward kinematics of the body, so that the body positions, the joint angles and the joint velocities of the reference stay consistent.
The joint angles receive, per joint group, a bias for the episode, two Ornstein--Uhlenbeck processes with correlation times of 0.7 and 0.08\,s, white noise at 30\,Hz, short bursts on one limb and rare implausible poses.
The root drifts horizontally by two Ornstein--Uhlenbeck processes with standard deviations of 0.04 and 0.18\,m and correlation times of 0.3 and 10\,s, and its heading drifts with a standard deviation of 11$^\circ$.
The reference also stays at an earlier frame for short periods.
At a perturbation scale of 1, the root-relative errors of the tracked bodies range from 45\,mm at the shank to 235\,mm at the hand.
The perturbation is a deterministic function of an episode identifier and the motion time, so that the policy observes the same perturbed reference in the simulator and in the imagined rollouts.

\begin{figure}[h]
  \centering
  \includegraphics[width=0.72\linewidth]{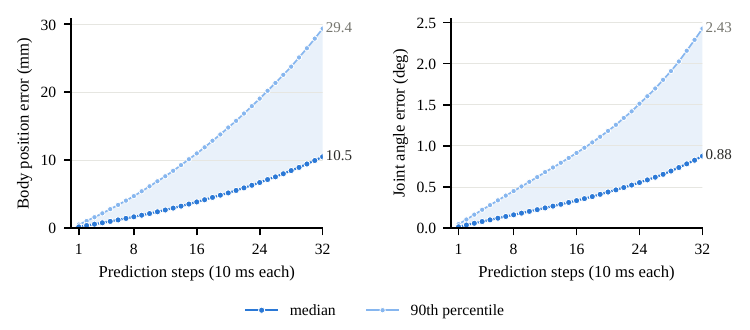}
  \caption{\textbf{Open-loop accuracy of the world model} of seed 0 against the number of predicted steps, from 12{,}636 recorded states of the policy on 300 BONES-SEED test motions with the recorded actions, as the median and 90th percentile over the states of the mean error of the 15 tracked bodies in the world frame and of the 31 joint angles. For comparison, the tracking error $E_\text{g-mpjpe}$ of the policy on the BONES-SEED test subjects is 58.0\,mm (\cref{tab:main}).}
  \label{fig:wmacc}
\end{figure}

\paragraph{World model.}
For the flat-ground stage, the input $z \in \mathbb{R}^{249}$ of $M_\psi$ consists of the root height, the root tilt (6), the root linear velocity in the heading frame and angular velocity in the root frame (6), the joint positions and velocities (62) and the positions relative to the root and velocities of 29 bodies (174), and the joint torque $\tau$ (31) is appended to it.
The output $\Delta x \in \mathbb{R}^{74}$ is the change of the root position and rotation (6), the joint positions (31), the root velocities (6) and the joint velocities (31) over one control step, with the root position and linear velocity in the heading frame and the root rotation and angular velocity in the root frame. Inputs and outputs are normalized with running statistics, and the output layer is initialized at zero.
The loss of $M_\psi$ is the squared error of the normalized state change averaged over windows of 8 steps that are predicted autoregressively from their first state, and the policy update starts after 300 world-model updates.
\Cref{fig:wmacc} gives the open-loop error of the world model against the number of predicted steps.
\paragraph{Geometry surrogate.}
It maps the 31 independent joint angles, scaled to $[-1, 1]$ by the joint limits, to the 354 muscle lengths and to the moment arms through a multilayer perceptron with three hidden layers of 512 units.
It predicts as a separate output the 1{,}038 moment arms that exceed $10^{-5}$\,m in at least one posture of the fitting set, and the other moment arms are set to zero.
The fitting set contains 1.5 million postures.
Half of them are poses of 500 training clips with Gaussian noise of standard deviation 0, 0.05, 0.15 or 0.3\,rad, and the other half are drawn uniformly within the joint limits.
For every posture, the simulator gives the muscle lengths and the moment arms with its tendon wrapping.
The loss is the squared error of the normalized lengths and moment arms, minimized with Adam for 60{,}000 steps of 8{,}192 postures, and the surrogate is kept fixed afterwards.

\paragraph{Reward.}
The reward of a step is $r = r_{\mathrm{task}} - 0.1\, \overline{u^2}$, where an overline denotes the mean over the muscles, the joints or the coordinates of the tracked bodies. The task reward is $r_{\mathrm{task}} = 0.5\, r_p + 0.05\, r_v + 0.4\, r_q + 0.05\, r_{\dot q} - 3\, \overline{(y_t - y_{t-1})^2} - 3\times10^{-5}\, \overline{\ddot q^2}$, where $r_p = \exp(-200\, \overline{\Delta p^2})$ and $r_v = \exp(-5\, \overline{\Delta v^2})$ compare the root-relative positions and velocities of the tracked bodies, $r_q = \exp(-20\, \overline{(q - q^{\mathrm{ref}})^2})$ and $r_{\dot q} = \exp(-\overline{(\dot q - \dot q^{\mathrm{ref}})^2})$ the joint positions and velocities, and $y$ is the raw policy output.

\section{Hyperparameters}
\label{app:hyper}

\Cref{tab:hyper} lists the settings of the final runs.
They are identical for the three seeds and, apart from the setting under study, for the ablations, and the reward weights are given in \cref{app:env}.

\begin{table}[h]
  \centering
  \caption{\textbf{Hyperparameters of the final runs.}}
  \label{tab:hyper}
  \small
  \resizebox{\linewidth}{!}{%
  \begin{tabular}{llr}
    \toprule
    Group & Setting & Value \\
    \midrule
    Simulation & time step, physics steps per control step, control rate & 2\,ms, 5, 100\,Hz \\
               & solver & Newton, 100 iterations, 50 line-search iterations \\
               & parallel environments & 4{,}096 \\
    Collection & control steps per iteration & 2 \\
               & replay buffer & 600 steps per environment \\
               & exploration noise (standard deviation), fraction of noise-free environments & 0.1, 0.25 \\
    World model & network $M_\psi$ & $3\times4{,}096$, LayerNorm, SiLU \\
               & window length $k$, minibatches $\times$ windows per iteration & 8, $8\times2{,}048$ \\
               & world-model updates before the first policy update & 300 \\
               & geometry surrogate & $3\times512$, SiLU, fixed \\
    Policy     & network & 2{,}048, 2{,}048, 1{,}024, LayerNorm, ELU, $\tanh$ output \\
               & initial excitation $u_0$ (initial policy excitation and first previous excitation) & 0.2 \\
               & reference steps ahead (control steps) & 1, 2, 5, 10, 20, 40 \\
               & imagination horizon $H$, rollouts per update & 32, 4{,}096 \\
               & per-sample state-gradient limit & $10\times$ batch median \\
               & policy gradient norm limit, element limit & 1, 0.1 \\
    Critic     & network & $3\times1{,}024$ \\
               & $\gamma$, $\lambda$, epochs $\times$ minibatches & 0.99, 0.95, $5\times4$ \\
    Task       & termination distance $d_{\max}$, smoothing width & 0.3\,m, 0.02\,m \\
    Terrain stage & heightmap points, spacing & $11\times11$, 0.1\,m \\
               & environments on terrain clips & 1{,}024 of 4{,}096 \\
               & world-model updates before the first policy update & 100 \\
               & episodes with a perturbed reference, scale of the perturbation & 50\,\%, 0.25 to 1.25 \\
    Optimizer  & Adam learning rate of all networks & $2\times10^{-4}$, then $6\times10^{-5}$ \\
    Software   & mjlab, MuJoCo and MuJoCo Warp, Warp & 1.0.0, 3.11.0, 1.14.0 \\
               & PyTorch, rsl\_rl~\cite{schwarke2025rslrl} & 2.10.0, 3.1.0 \\
    \bottomrule
  \end{tabular}}
\end{table}

\section{Training stages, stopping and checkpoint selection}
\label{app:stages}

Each training stage stops according to a validation metric, with rules fixed before the final runs (\cref{tab:stages}).
Every 500 policy updates, the current checkpoint is evaluated on the validation set, a fixed subset of 512 clips of the 10 BONES-SEED validation subjects.
The evaluation runs once with the deterministic policy and once with action noise of standard deviation 0.1, and the success rate under the 0.3\,m rule is recorded for both.
A plateau is declared when, for both success rates, the median of the last four validated checkpoints is at most one percentage point above the largest median reached at least 2{,}000 policy updates earlier. A stage needs at least eight validated checkpoints and 2{,}000 policy updates before a plateau can be declared.
At the first plateau the learning rate of all networks is lowered once from $2\times10^{-4}$ to $6\times10^{-5}$, and at the second plateau the training stops.
The checkpoint is then selected among all validated checkpoints by a rule that does not depend on their order.
A checkpoint is eligible when both of its success rates lie within $\max(2, 2\,\mathrm{SE})$ percentage points of the highest value of that rate, with the binomial standard error SE on the validation set.
Among the eligible checkpoints, the one with the highest mean of the two rates is chosen, and ties are resolved by the lower mean excitation.
The test sets are used for neither the transitions nor the selection.

\begin{table}[h]
  \centering
  \caption{\textbf{Training stages.} Every transition is triggered by the plateau rule on the validation set, and the checkpoint is chosen by the selection rule, both fixed before the final runs.}
  \label{tab:stages}
  \small
  \begin{tabular}{p{0.15\linewidth}p{0.38\linewidth}p{0.38\linewidth}}
    \toprule
    Stage & Purpose & Start and end \\
    \midrule
    1a, flat ground & learn to track the flat-ground motions & starts from random weights after 300 world-model updates, ends at the first plateau \\
    1b, flat ground & refine the policy at the lower learning rate & starts at the first plateau, ends at the second plateau, followed by the checkpoint selection \\
    2, terrain & extend the policy to stairs, ramps, boxes and seats & starts from the checkpoint selected in stage 1 with the heightmap branches at zero, lowers the learning rate at the first plateau of the mean success on the flat-ground and terrain validation sets and ends at the second, followed by the checkpoint selection \\
    \bottomrule
  \end{tabular}
\end{table}

\section{Corrections of the body model}
\label{app:body}
The five corrections of \cref{sec:env} change the inertial and muscle-path data of the released MyoFullBody model and leave its joints, frames, joint ranges and markers unchanged.
Each addresses an issue reported in the issue tracker of the MyoSuite models\footnote{\url{https://github.com/MyoHub/myo_sim}, issues and pull requests 128, 138, 144, 154, 155 and 163.}.
1) The placeholder rotational inertia of 1\,kg\,m$^2$ on eight auxiliary shoulder bodies of 1\,g inflates the trunk and shoulder torques in fast arm motions and is set to $10^{-2}$\,kg\,m$^2$.
2) The pelvis mass is counted twice, once in the pelvis of the leg model and once in the sacrum of the torso model, and the sacrum is made massless.
3) The head mass is raised from 2.13 to 5.85\,kg, which is the head-and-neck fraction of 6.94\,\% of \citet{deleva1996adjustments} applied to the body mass of 84.27\,kg before the corrections, and the trunk mass is lowered by the same amount from 18.62 to 14.90\,kg.
4) The path of the left T12 longissimus is mirrored from the right side.
5) The elbow wrap of the long head of the biceps is removed from its path.
The total mass changes from 84.27 to 76.78\,kg, and the muscle-geometry surrogate is fitted to the corrected body.

\section{Quality filter}
\label{app:quality}
\Cref{tab:quality} lists the criteria of the quality filter. Clips that meet any criterion are removed, except for the 108 KIT benchmark test motions retained as described below.
The thresholds were fixed before training, based on the checks of the retargeting pipeline and on physical limits.
The filter removes 6.1\,\% of the hours of the BONES-SEED flat-ground training source, 0.9 and 2.6\,\% of its box and seat motions, 4.1\,\% of Gait120, 4.4\,\% of the Vielemeyer ramps and 35\,\% of the KIT terrain motions.
Five of the 108 KIT test motions meet the criterion on the median marker distance, with 10.1 to 12.6\,mm, and one of them also meets the criterion on the 95th percentile. All 108 motions are kept for testing.

\begin{table}[h]
  \centering
  \caption{\textbf{Criteria of the quality filter.} A clip is removed when it meets any criterion.}
  \label{tab:quality}
  \small
  \resizebox{\linewidth}{!}{%
  \begin{tabular}{ll}
    \toprule
    Group & Criterion \\
    \midrule
    Marker fit & median body-marker distance $>$ 10\,mm, or 95th percentile of the distances $>$ 40\,mm \\
    Solver & at least two markers of a limb $>$ 6\,cm away for more than 0.2\,s, or frames where the joint limits prevent the marker fit for more than 0.2\,s \\
    Feet & airborne phase longer than 1\,s or not ballistic within 2\,cm \\
         & stance with a median foot height above the support $>$ 2\,cm \\
         & sole or hand $>$ 2\,cm, or another body $>$ 5\,cm below the support in any frame \\
         & a supporting foot slides faster than 0.2\,m/s for at least 0.1\,s \\
    Physics & an independent joint more than 1$^\circ$ outside its range \\
         & joint speed $>$ 22.5\,rad/s, or joint acceleration $>$ 848\,rad/s$^2$ \\
         & root speed $>$ 10\,m/s, root acceleration $>$ 100\,m/s$^2$ or root angular speed $>$ 40\,rad/s \\
         & a forearm, hand or leg more than 6\,cm inside a non-adjacent body \\
    Length & shorter than 1.12\,s, the sum of the warm-up, the imagination horizon and the steps ahead of the reference \\
    \bottomrule
  \end{tabular}}
\end{table}

\section{Detailed test results}
\label{app:detail}
\Cref{tab:categories} gives the success of the flat-ground policy on the BONES-SEED test subjects per motion category, and \cref{tab:boxes} gives the success of the terrain policy on the two kinds of box motions.

\begin{table}[h]
  \centering
  \caption{\textbf{Success per motion category} on the BONES-SEED test subjects with the deterministic flat-ground policy, mean $\pm$ standard deviation over three seeds, in \%.}
  \label{tab:categories}
  \small
  \begin{tabular}{lcccc}
    \toprule
    Category & Clips & 0.5\,m & 0.3\,m & 0.15\,m \\
    \midrule
    Dancing & 798 & 96.2 $\pm$ 1.0 & 94.7 $\pm$ 0.8 & 81.3 $\pm$ 1.3 \\
    Locomotion, styles & 347 & 99.8 $\pm$ 0.3 & 99.3 $\pm$ 0.6 & 88.7 $\pm$ 2.2 \\
    Baseline & 256 & 99.0 $\pm$ 0.2 & 95.8 $\pm$ 0.5 & 85.7 $\pm$ 1.0 \\
    Locomotion, neutral & 130 & 100.0 $\pm$ 0.0 & 100.0 $\pm$ 0.0 & 97.4 $\pm$ 4.4 \\
    Gestures & 21 & 100.0 $\pm$ 0.0 & 100.0 $\pm$ 0.0 & 90.5 $\pm$ 0.0 \\
    Stunts & 15 & 100.0 $\pm$ 0.0 & 100.0 $\pm$ 0.0 & 97.8 $\pm$ 3.8 \\
    Complex actions & 7 & 100.0 $\pm$ 0.0 & 100.0 $\pm$ 0.0 & 52.4 $\pm$ 8.2 \\
    \bottomrule
  \end{tabular}
\end{table}

\begin{table}[h]
  \centering
  \caption{\textbf{Box motions} of the BONES-SEED test subjects with the terrain policy, mean $\pm$ standard deviation over three seeds, in \%, for the deterministic policy and with action noise of standard deviation 0.1.}
  \label{tab:boxes}
  \small
  \begin{tabular}{llcccc}
    \toprule
    Motions & Policy & Clips & 0.5\,m & 0.3\,m & 0.15\,m \\
    \midrule
    Steps onto and off a 50\,cm box & deterministic & 44 & 94.7 $\pm$ 5.2 & 91.7 $\pm$ 3.5 & 83.3 $\pm$ 3.5 \\
     & action noise & 44 & 77.3 $\pm$ 9.1 & 75.8 $\pm$ 9.2 & 61.4 $\pm$ 9.1 \\
    Jumps onto, off and over boxes & deterministic & 97 & 35.1 $\pm$ 3.6 & 30.9 $\pm$ 1.8 & 16.8 $\pm$ 1.6 \\
     & action noise & 97 & 31.3 $\pm$ 1.6 & 29.9 $\pm$ 0.0 & 14.1 $\pm$ 0.6 \\
    \bottomrule
  \end{tabular}
\end{table}

\section{Kinetics in walking}
\label{app:kinetics}
The simulated gait cycle of each trial is aligned at the heel strike of the right foot, where its ground contact force exceeds 20\,N. As in the EMG comparison, the first 0.5\,s and the last 0.4\,s of each trial are excluded, and the trials with a detected heel strike and a complete gait cycle are 347, 354 and 353 of the 538 clips for the three seeds, from 102 subjects.
The simulated ground reaction force is the whole-body force $m(a_\text{com}-g)$ from the acceleration of the center of mass, divided between the feet in proportion to their ground contact forces, and the joint torques of the muscles are the actuator forces of the simulator mapped to the joints.
The center of mass and the joint torques of the simulation and the marker, force-plate and moment data of \citet{schreiber2019multimodal} and \citet{camargo2021comprehensive} are low-pass filtered at 6\,Hz with a fourth-order zero-lag Butterworth filter.
Each dataset is cut at its own heel strikes, resampled to 101 points per stride and averaged per subject and then across subjects.
The data of \citet{schreiber2019multimodal} are restricted to 45 subjects walking at 1.13 to 1.45\,m/s, the 10th to 90th percentiles of the speeds in Gait120, and their joint moments are computed by inverse dynamics.
\citet{fukuchi2018public} contribute the published curves of 42 subjects at comfortable speed, and \citet{camargo2021comprehensive} the joint moments of 22 subjects in level-ground walking at self-selected speeds that are not matched to Gait120.
Forces are normalized by body weight and moments by body mass, using the model mass of 76.78\,kg for the simulation.

\section{Results with action noise}
\label{app:noise}
\Cref{tab:noise,tab:noise_terrain} repeat the tests of \cref{tab:main,tab:terrain} with Gaussian action noise of standard deviation 0.1 on the policy output.

\begin{table}[h]
  \centering
  \caption{\textbf{Tracking on flat ground with action noise}, mean $\pm$ standard deviation over three seeds. Success rates in \%, $E_\text{g-mpjpe}$ in mm over all frames.}
  \label{tab:noise}
  \small
  \begin{tabular}{lccccc}
    \toprule
    Test set & 0.5\,m & 0.3\,m & 0.15\,m & No-fall & $E_\text{g-mpjpe}$ \\
    \midrule
    BONES-SEED, test subjects & 96.8 $\pm$ 0.5 & 94.4 $\pm$ 0.6 & 79.4 $\pm$ 1.6 & 97.2 $\pm$ 0.4 & 69.7 $\pm$ 5.7 \\
    KIT, 108 test motions & 100.0 $\pm$ 0.0 & 99.7 $\pm$ 0.5 & 89.8 $\pm$ 1.6 & 100.0 $\pm$ 0.0 & 47.6 $\pm$ 6.9 \\
    BONES-SEED, training subset & 96.5 $\pm$ 0.4 & 95.7 $\pm$ 0.4 & 86.5 $\pm$ 2.7 & 97.2 $\pm$ 0.3 & 64.6 $\pm$ 5.6 \\
    \bottomrule
  \end{tabular}
\end{table}

\begin{table}[h]
  \centering
  \caption{\textbf{Tracking on terrain with action noise} for the terrain policy, mean $\pm$ standard deviation over three seeds. Success rates in \%, and MPJPE in mm over the frames before the first 0.3\,m failure as in \cref{tab:terrain}.}
  \label{tab:noise_terrain}
  \small
  \begin{tabular}{lccccc}
    \toprule
    Test set & Clips & 0.5\,m & 0.3\,m & 0.15\,m & MPJPE \\
    \midrule
    Gait120, stairs and ramps & 167 & 90.8 $\pm$ 0.9 & 90.0 $\pm$ 0.9 & 84.0 $\pm$ 0.7 & 65.8 $\pm$ 2.3 \\
    Vielemeyer, ramps & 95 & 100.0 $\pm$ 0.0 & 100.0 $\pm$ 0.0 & 89.8 $\pm$ 15.8 & 61.5 $\pm$ 5.5 \\
    BONES-SEED seats & 51 & 98.0 $\pm$ 0.0 & 94.8 $\pm$ 3.0 & 81.7 $\pm$ 1.1 & 46.8 $\pm$ 2.8 \\
    BONES-SEED boxes & 141 & 45.6 $\pm$ 1.8 & 44.2 $\pm$ 2.9 & 28.8 $\pm$ 2.5 & 66.8 $\pm$ 1.2 \\
    KIT terrain & 156 & 76.9 $\pm$ 2.8 & 70.5 $\pm$ 5.1 & 60.3 $\pm$ 2.2 & 59.6 $\pm$ 3.5 \\
    \midrule
    BONES-SEED flat ground & 1{,}574 & 97.6 $\pm$ 0.4 & 94.5 $\pm$ 0.2 & 75.6 $\pm$ 0.2 & 56.5 $\pm$ 1.3 \\
    KIT, 108 test motions & 108 & 99.7 $\pm$ 0.5 & 98.5 $\pm$ 0.5 & 86.7 $\pm$ 3.9 & 47.9 $\pm$ 6.1 \\
    \bottomrule
  \end{tabular}
\end{table}

\section{Push test}
\label{app:push}
The push is a constant horizontal force on the pelvis, applied at its center of mass for 0.2\,s, with impulses from 0 to 250\,N\,s in steps of 10\,N\,s (\cref{tab:push}).
Its start is drawn uniformly between 1\,s after the start and 1\,s before the end of the clip, or set to the middle of clips shorter than 2.5\,s, and its direction is drawn uniformly in the horizontal plane, both fixed for each clip.
For each seed, the impulse at which half of the motions fall is interpolated linearly between the levels. For push directions within $\pm 45^\circ$ of each axis of the pelvis heading, this impulse is $126 \pm 6$\,N\,s from behind, $116 \pm 11$\,N\,s from the front, $134 \pm 11$\,N\,s to the left and $128 \pm 7$\,N\,s to the right.

\begin{table}[!htb]
  \centering
  \caption{\textbf{Push test} on 300 BONES-SEED test motions with the deterministic policy, mean $\pm$ standard deviation over three seeds, in \%.}
  \label{tab:push}
  \small
  \begin{tabular}{rrccc}
    \toprule
    Impulse (N\,s) & Force (N) & Success 0.5\,m & Success 0.3\,m & No-fall \\
    \midrule
    0 & 0 & 98.9 $\pm$ 0.4 & 97.8 $\pm$ 0.5 & 98.9 $\pm$ 0.4 \\
    40 & 200 & 97.9 $\pm$ 0.2 & 94.9 $\pm$ 0.4 & 98.1 $\pm$ 0.5 \\
    60 & 300 & 95.2 $\pm$ 0.4 & 84.0 $\pm$ 2.6 & 96.2 $\pm$ 0.5 \\
    80 & 400 & 85.2 $\pm$ 3.3 & 52.4 $\pm$ 3.2 & 91.4 $\pm$ 2.2 \\
    100 & 500 & 56.7 $\pm$ 8.4 & 21.0 $\pm$ 2.7 & 76.0 $\pm$ 6.0 \\
    120 & 600 & 27.6 $\pm$ 4.7 & 3.6 $\pm$ 1.7 & 55.6 $\pm$ 7.4 \\
    150 & 750 & 4.3 $\pm$ 1.0 & 0.8 $\pm$ 0.4 & 28.7 $\pm$ 5.3 \\
    200 & 1{,}000 & 0.4 $\pm$ 0.2 & 0.0 $\pm$ 0.0 & 8.4 $\pm$ 1.2 \\
    250 & 1{,}250 & 0.0 $\pm$ 0.0 & 0.0 $\pm$ 0.0 & 5.6 $\pm$ 0.5 \\
    \bottomrule
  \end{tabular}
\end{table}

\section{Compute}
\label{app:compute}
All Myodyssey training times are measured on one NVIDIA RTX PRO 6000 Blackwell Max-Q GPU, excluding validation evaluations.
A training iteration takes 1.31\,s, and the world model is fitted for 214\,s before the first policy update. 
We take the time of the checkpoint that the selection rule chooses as the convergence time.
The flat-ground policy converges after 3.3, 3.5 and 4.1\,h for the three seeds, about 3.6\,h on average.
Early stopping ends the training after 4.1, 4.1 and 4.4\,h, later than the selected checkpoints, because the plateau rule of \cref{app:stages} compares the median success of the last four validated checkpoints with the largest median reached at least 2{,}000 policy updates earlier.
The terrain stage takes 1.69\,s per iteration and reaches its selected checkpoint after 5.4, 5.4 and 6.6\,h and early stopping after 6.7, 6.5 and 7.3\,h.
\Cref{tab:compute} lists the simulation steps, the steps imagined in the world model and the policy updates of the three runs until early stopping, and the selected checkpoints correspond to 76.3, 80.4 and 92.7 million simulation steps.
\Cref{tab:train_cost} lists training costs reported by related muscle-driven trackers together with the training budgets evaluated in this study.

\begin{table}[h]
  \centering
  \caption{\textbf{Flat-ground training until early stopping} for the three seeds of the main configuration. The terrain stage is not included.}
  \label{tab:compute}
  \small
  \begin{tabular}{lccc}
    \toprule
     & Seed 0 & Seed 1 & Seed 2 \\
    \midrule
    Simulation steps & $93.2\times10^{6}$ & $93.2\times10^{6}$ & $101.4\times10^{6}$ \\
    Imagined steps & $1.45\times10^{9}$ & $1.45\times10^{9}$ & $1.58\times10^{9}$ \\
    Policy updates & 11{,}057 & 11{,}054 & 12{,}053 \\
    \bottomrule
  \end{tabular}
\end{table}

\begin{table}[h]
  \centering
  \caption{\textbf{Reported training cost of muscle-driven motion trackers}, with different data, hardware and success rules. --: not stated. $^e$Estimated from the number of KIT clips and the mean clip length of the KINESIS set. $^m$Including left--right mirrored copies. $^q$10 days as quoted by MuscleMimic. $^u$Four experts and fine-tuning, without the distillation. $^d$About 20\,h per $10^9$ steps for the steps of the released checkpoint. $^g$About 20\,h on four GPUs. $^r$For MyoFullBody, 1\,h of training with the fast recipe in the paper~\cite{he2026musclesupermimic}, after which its root-relative MPJPE on the KIT test clips is about 25\,mm. $^i$With $3.4\times10^9$ further steps imagined in the world model. $^s$Flat-ground and terrain stages, each until early stopping, about 11\,h in total. The root-relative MPJPE of the flat-ground policy on the KIT test clips is 19.4\,mm (\cref{tab:main}).}
  \label{tab:train_cost}
  \small
  \resizebox{\linewidth}{!}{%
  \begin{tabular}{llcllcc}
    \toprule
    Method & Learner & Motion (h) & Hardware & Environments & Simulator steps & GPU hours \\
    \midrule
    KINESIS~\cite{simos2026kinesis} & PPO & 1.8 & 1 A100, 128 CPU threads & 128 & -- & 240$^q$ \\
    UniExo~\cite{yuan2026uniexo} & PPO & 1.2 & 1 RTX PRO 6000 & 8{,}192 & $2.9$--$4.9\times10^9$ per expert & 114--174$^u$ \\
    MuscleMimic~\cite{li2026musclemimic} & PPO & 1.5 & 1 H100 & 8{,}192 & $3.19\times10^9$ & 64$^d$ \\
    MyoAcross~\cite{wang2026myoacross} & PPO & 0.1$^e$ & 4 RTX 5090 & 8{,}192 & $3.93\times10^9$ & 168 \\
    TERRA~\cite{simos2026terra} & PPO & 9.4 & 4 A100 & 8{,}192 & $4\times10^9$ & 80$^g$ \\
    muscle-supermimic~\cite{he2026musclesupermimic} & BPTT in world model & 1.5 & 1 RTX 5090 & 1{,}024 & -- & 1$^r$ \\
    \midrule
    \rowcolor{oursrow} Myodyssey (ours) & BPTT in world model & 182.4$^m$ & 1 RTX PRO 6000 & 4{,}096 & $0.2\times10^9$\,$^i$ & 4.2 + 6.8$^s$ \\
    \bottomrule
  \end{tabular}}
\end{table}

\end{document}